\documentclass[10pt,twocolumn,letterpaper]{article}

\usepackage{cvpr}              % To produce the CAMERA-READY version
\usepackage[dvipsnames]{xcolor}

\usepackage{bm}
\usepackage{bbding}
\usepackage{multirow}
\newcommand\blfootnote[1]{%
  \begingroup
  \renewcommand\thefootnote{}\footnote{#1}%
  \addtocounter{footnote}{-1}%
  \endgroup
}

\definecolor{cvprblue}{rgb}{0.21,0.49,0.74}
\usepackage[pagebackref,breaklinks,colorlinks,citecolor=cvprblue]{hyperref}

\def\paperID{***} % *** Enter the Paper ID here
\def\confName{3DV\xspace}
\def\confYear{2026\xspace}

\title{UniGaussian: Driving Scene Reconstruction from \\ Multiple Camera Models via Unified Gaussian Representations}

\author{Yuan Ren\textsuperscript{1}, Guile Wu\textsuperscript{1}, Runhao Li\textsuperscript{1,2,}{\footnotemark[1]}, Zheyuan Yang\textsuperscript{1,2,}{\footnotemark[1]}, Yibo Liu\textsuperscript{1,3,}{\footnotemark[1]}, \\
Xingxin Chen\textsuperscript{1}, Tongtong Cao\textsuperscript{1}, and Bingbing Liu\textsuperscript{1,}{\footnotemark[2]}\\
\textsuperscript{1}Huawei Noah's Ark Lab \quad \textsuperscript{2}University of Toronto \quad \textsuperscript{3}York University\\
{\tt\small\{yuan.ren3, xingxin.chen1, caotongtong, liu.bingbing\}@huawei.com, guile.wu@outlook.com}\\
{\tt\small\{rain.li, andrewzheyuan.yang\}@mail.utoronto.ca, buaayorklau@gmail.com}
}

\begin{document}
% \maketitle
\twocolumn[{%
\renewcommand\twocolumn[1][]{#1}%
\maketitle
\vspace{-1cm}
\begin{center}
    \centering
    \captionsetup{type=figure}
    \includegraphics[width=0.75\textwidth]{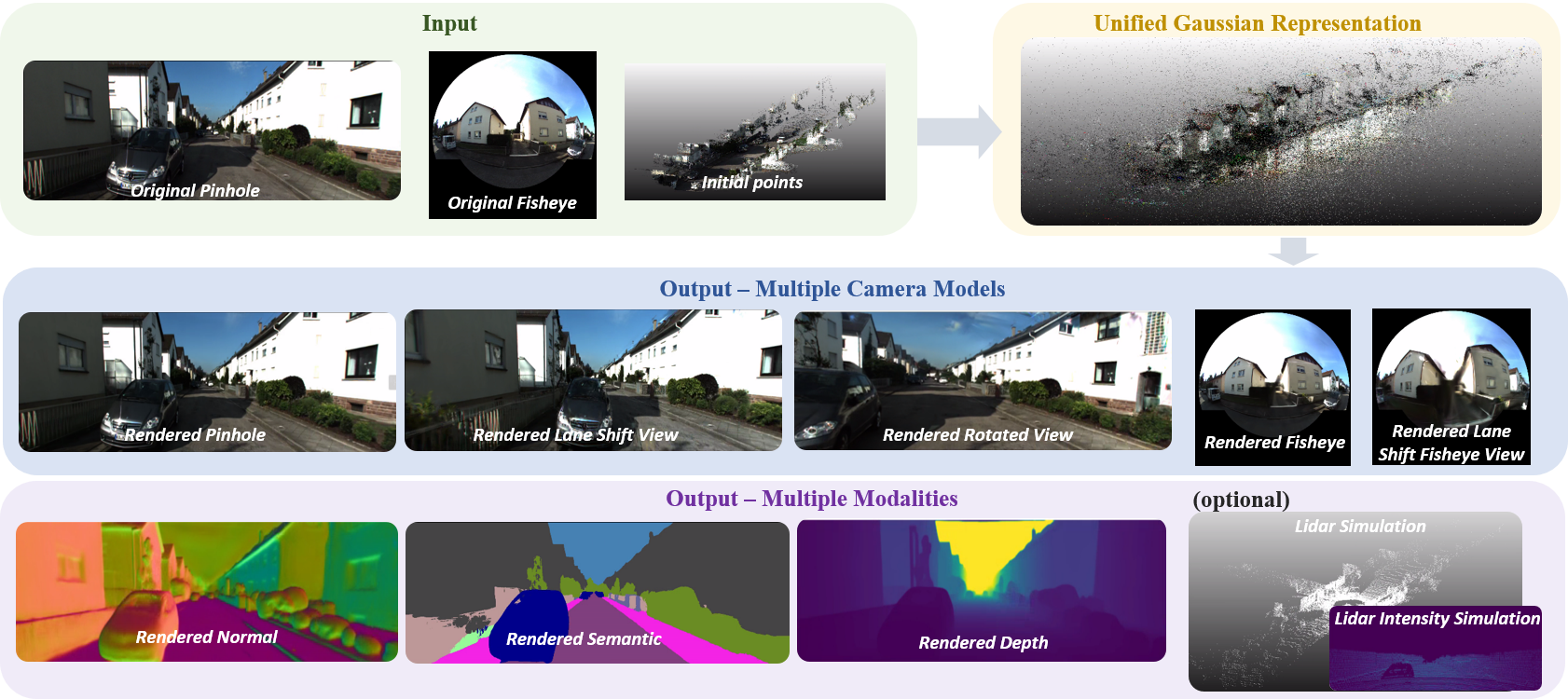}
    \vspace{-0.3cm}
    \captionof{figure}{
        An illustration of the proposed approach.
        UniGaussian reconstructs 3D driving scenes by learning unified 3D Gaussian representations from multiple input sources. It achieves holistic driving scene understanding and models multiple sensors (pinhole cameras and fisheye cameras) and modalities (semantic, normal, depth, and optional LiDAR point clouds).
        }
    \label{fig:illustration}
\end{center}%
}]

\blfootnote{
\textsuperscript{*}R. Li, Z. Yang, and Y. Liu made contributions during their internships at Huawei Canada.
\textsuperscript{$\dagger$}B. Liu is the corresponding author.\\
}

\begin{abstract}
Urban scene reconstruction is crucial for real-world autonomous driving simulators.
Although existing methods have achieved photorealistic reconstruction, they mostly focus on pinhole cameras and neglect fisheye cameras.
In fact, how to effectively simulate fisheye cameras in driving scenes remains an unsolved problem.
In this work, we propose UniGaussian, a novel approach that learns unified 3D Gaussian representations from multiple camera models for urban scene reconstruction in autonomous driving.
Our contributions are two-fold.
First, we propose a new differentiable rendering method that distorts 3D Gaussians using a series of affine transformations tailored to fisheye camera models.
This addresses the compatibility issue of 3D Gaussian splatting with fisheye cameras, which is hindered by light ray distortion caused by lenses or mirrors.
Besides, our method maintains real-time rendering while ensuring differentiability. 
Second, built on the differentiable rendering method, we design a new framework that learns unified Gaussian representations from multiple camera models.
By applying affine transformations to adapt different camera models and regularizing the shared Gaussians with supervision from different modalities, our framework learns unified 3D Gaussian representations with input data from multiple sources and achieves holistic driving scene understanding.
As a result, our approach models multiple sensors (pinhole and fisheye cameras) and modalities (depth, semantic, normal, and LiDAR point clouds). 
Our experiments show that our method achieves superior rendering quality and fast rendering speed for driving scene simulation.
\end{abstract}    
\section{Introduction}
\label{sec:intro}

Urban scene reconstruction aims to reconstruct real-world driving scenes from images and achieve photorealistic rendering~\cite{yang2023unisim,tao2024alignmif,zhou2024hugs,chen2023periodic}.
It is crucial for realizing real-world closed-loop evaluation in end-to-end autonomous driving simulators~\cite{yang2023unisim,yan2024forging}.
One of the most popular methods for driving scene reconstruction is the use of Neural Radiance Field (NeRF)~\cite{mildenhall2021nerf}, an implicit radiance field representation.
Although NeRF-based methods~\cite{yang2023unisim,tao2024alignmif,xie2023snerf,guo2023streetsurf,ost2021neural} have achieved photorealistic rendering, they suffer from slow rendering speed and expensive training cost.
Recently, some researchers have resorted to 3D Gaussian Splatting (3DGS)~\cite{kerbl3Dgaussians}, an explicit radiance field representation, for urban scene reconstruction.
3DGS-based methods~\cite{zhou2024drivinggaussian,zhou2024hugs,chen2023periodic,yan2024street} use explicit 3D Gaussian primitives to represent 3D scenes and employ a differentiable tile rasterizer for rendering.
This enhances the editability of driving scenes and achieves real-time photorealistic rendering.

However, existing driving scene reconstruction methods mainly focus on rendering from pinhole cameras and largely neglect fisheye cameras.
Compared with pinhole cameras, fisheye cameras provide a wider field of view (FOV) essential for navigation and perception tasks in autonomous driving, especially for near-field sensing and automatic parking~\cite{rashed2021generalized,Liao2022PAMI,li2019distortion}.
Nevertheless, it is non-trivial to adapt 3DGS to fisheye cameras.
For example, the direct application of 3DGS to fisheye cameras is impeded by distortions induced by camera optics, which disrupt the affine transformations of 3D Gaussians.
Intuitively, an engineering strategy to tackle this issue is directly rectifying fisheye images and treating them as pinhole images for model training.
However, this strategy yields suboptimal rendering quality, particularly in regions with a large field of view angle.
Another strategy is to render multiple cubic images and subsequently amalgamate them to construct fisheye images~\cite{komatsu2020360,wang2020bifuse}.
Nevertheless, this strategy introduces noticeable artifacts indicative of the stitching process.
In fact, how to effectively render driving scenes from fisheye images for simulation remains an open question.

\begin{figure*}[tb]
  \centering
   \includegraphics[width=0.9\textwidth]{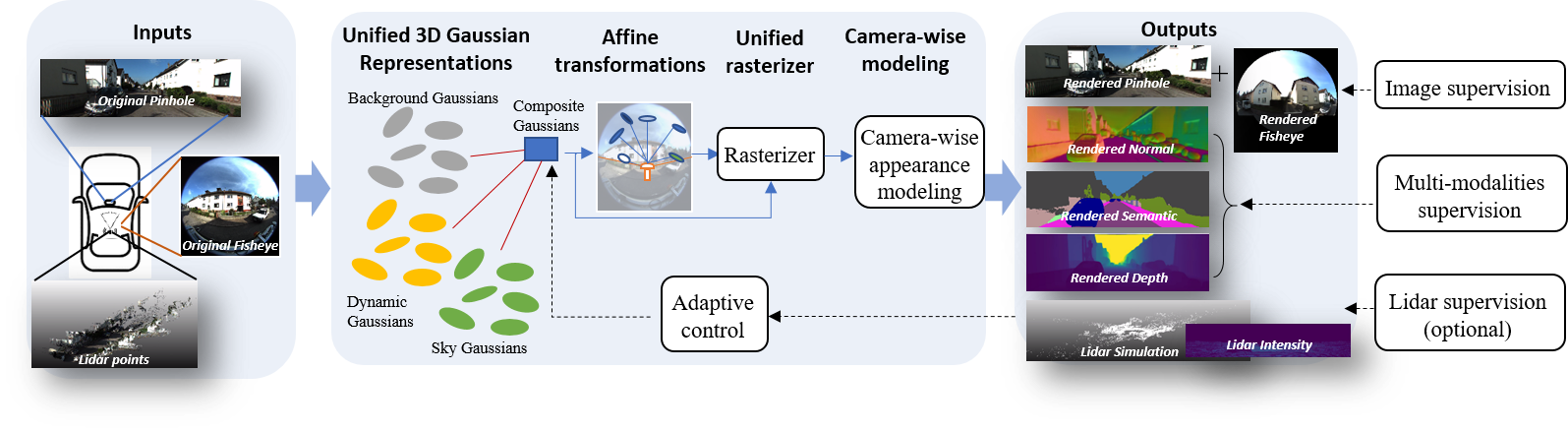}
   \vspace{-0.6cm}
   \caption{The framework of our UniGaussian approach to driving scene reconstruction with multiple camera models.
   Our approach achieves holistic driving scene understanding by modeling multiple sensors and modalities.}
   \label{fig:framework}
\end{figure*}

To address the aforementioned challenges, this work presents UniGaussian, a novel 3DGS-based approach that learns unified Gaussian representations from multiple camera models for urban scene reconstruction in autonomous driving.
First, we aim to address the compatibility issue of 3DGS with fisheye cameras.
To this end, we propose a novel differentiable rendering method for 3DGS tailored to fisheye cameras.
In our method, the distortion of light rays caused by fisheye cameras is translated into the deformation of the radiance field.
This is realized through a series of affine transformations, including translation, rotation, and stretching, applied to 3D Gaussians.
These operations maintain the parallelization strategy and real-time performance of 3DGS, while ensuring differentiability.
Second, with the proposed differentiable rendering method, we design a new framework that jointly optimizes unified 3D Gaussian representations from multiple camera models for driving scene reconstruction.
We apply affine transformations on 3D Gaussians to adapt them for different camera models and regularize the shared Gaussians with supervision from different modalities.
In this way, our framework learns unified 3D Gaussian representations with input data from multiple sources and achieves holistic driving scene understanding.
As illustrated in \cref{fig:illustration}, our approach models multiple sensors, including pinhole cameras and fisheye cameras, and multiple modalities, including depth, semantic, normal, and optional LiDAR point clouds.
Besides, since we use a unified rasterizer for 3D Gaussian rendering, our framework shows adaptability to various differentiable camera models and maintains real-time performance.
The framework of the proposed approach is depicted in \cref{fig:framework}.

In summary, our \textbf{contributions} are two-fold:
\begin{itemize}[left=1.2em]
  \item {We propose a novel differentiable rendering method for 3DGS tailored to driving scene reconstruction from fisheye cameras. Through a series of affine transformations, our method has adaptability to various fisheye camera models for driving scene reconstruction with large FOVs, does not significantly increase GPU memory consumption and maintains real-time rendering.}
  \item {Built on the proposed differentiable rendering method, we propose to learn unified Gaussian representations from multiple camera models for driving scene reconstruction.
  Our approach achieves holistic driving scene understanding by modeling multiple sensors and modalities.
  To the best of our knowledge, no existing driving scene reconstruction method simulates both pinhole and fisheye cameras in a unified framework.}
\end{itemize}

\section{Related Work}
\label{sec:related_work}

\paragraph{3D Gaussian Splatting.}
\label{sec:related_work_3dgs}
3DGS~\cite{kerbl3Dgaussians} is a real-time radiance field rendering approach.
It represents a 3D scene with a set of explicit 3D Gaussian primitives.
Each primitive is defined by a position (mean) $\mu\in\mathbb{R}^3$, a covariance matrix $\Sigma\in\mathbb{R}^{3{\times}3}$, an opacity $o\in\mathbb{R}^1$ and spherical harmonics (colors).
To render images, these 3D Gaussian primitives are projected from 3D space onto 2D image plane to compute the color for each pixel with a differentiable tile-based rasterizer.
3DGS achieves highly effective training and photorealistic real-time rendering.  
We refer readers to \cite{kerbl3Dgaussians} for more details of 3DGS.

With the incredible success achieved by 3DGS, many 3DGS-based methods have emerged with the aim of solving the problems of anti-aliasing~\cite{yu2024mip}, acceleration~\cite{fan2023lightgaussian}, relighting~\cite{gao2023relightable}, sparse-view synthesis~\cite{xiong2023sparsegs}, 4D reconstruction~\cite{wu20244d}, \etc.
Most of these works focus on reconstruction with pinhole camera models only.
Recently, Liao~\etal~\cite{liao2024fisheye} adapt 3DGS to fisheye cameras by recalculating 3DGS projection and gradients, but as they point out, their method is only based on ideal camera models with equidistant projection and not for generic fisheye camera models and real large-FOV cameras.
This hinders their use in driving scene reconstruction because fisheye cameras in driving scene are usually generic models and have large FOVs.
In contrast, our work proposes a new differentiable rendering method for 3DGS tailored to driving scene reconstruction with fisheye cameras.
Through a series of affine transformations, our method has adaptability to various fisheye camera models for driving scene reconstruction with large FOVs, does not significantly increase GPU memory consumption and maintains real-time rendering.

\vspace{-0.3cm}
\paragraph{Driving Scene Reconstruction.}
\label{sec:related_work_driving}
Typically, autonomous driving scene simulation involves driving scene reconstruction~\cite{yang2023unisim,xie2023snerf,tao2024alignmif} and road asset reconstruction~\cite{liu2023mv,liu2024vqa}.
Our work mainly focuses on driving scene reconstruction from images.
Contemporary driving scene reconstruction methods can be mainly categorized into two types, namely NeRF-based methods and 3DGS-based methods.
Generally, NeRF-based methods~\cite{tao2024alignmif,xie2023snerf,guo2023streetsurf,ost2021neural,yang2023unisim} perform dense ray sampling and leverage a multi-layer perception with many fully-connected layers to represent 3D scenes.
Although they have shown promising results for photorealistic rendering, they are usually slow for both training and rendering.
This hinders their application to real-world autonomous driving simulators.
On the other hand, 3DGS-based methods~\cite{zhou2024drivinggaussian,zhou2024hugs,chen2023periodic,yan2024street} appear to be more efficient for driving scene reconstruction due to the efficiency of explicit 3D Gaussian representations and differentiable rasterization.
They are able to achieve photorealistic reconstruction and real-time rendering.
Nevertheless, existing NeRF-based and 3DGS-based methods neglect the importance of fisheye camera simulation for autonomous driving.
In autonomous driving, fisheye cameras are useful for providing a wider field of view, especially for near-field sensing and automatic parking.
Our work differs from existing works in that the proposed method is built on a novel differentiable rendering method tailored to fisheye cameras and the proposed framework learns unified 3D Gaussians by modeling multiple sensors and modalities.

Although we use composite Gaussians as the driving scene representation following~\cite{zhou2024hugs,zhou2024drivinggaussian,yan2024street}, our approach significantly differs from~\cite{zhou2024hugs,zhou2024drivinggaussian,yan2024street}.
Our approach presents a novel differentiable rendering method tailored to fisheye cameras and constructs a unified framework to learn unified 3D Gaussians by modeling multiple sensors and modalities.
In contrast, \cite{zhou2024hugs,zhou2024drivinggaussian,yan2024street} are designed to model driving scene with pinhole cameras only.
Specifically, \cite{yan2024street} constructs composite street Gaussians as driving scene representation and employs tracking pose optimization to improve foreground object modeling;
\cite{zhou2024hugs} employs composite Gaussians for holistic scene understanding with a unicycle model for foreground object modeling;
\cite{zhou2024drivinggaussian} presents incremental 3D Gaussians and dynamic Gaussian graph for driving scene reconstruction.
None of them simulates multiple camera models in a unified framework for driving scene reconstruction.
Note that post-processing or redistrotion is not a practical solution to 3DGS fisheye rendering in driving scene reconstruction, which yields suboptimal rendering quality in regions with a large FOV (see \cref{sec:exp_fisheye_simulation} for comparison with 3DGS+Undistort) and prevents each camera model from learning complementary information.

\section{Methodology}
\label{sec:methodology}

This work focuses on driving scene reconstruction from multiple camera models, including pinhole and fisheye cameras. 
Before delving into driving scene reconstruction from multiple camera models, it is necessary to discuss differentiable rendering of 3DGS for fisheye cameras.
Thus, in \cref{sec:method_math}, we briefly introduce mathematical models of fisheye cameras;
then, in \cref{sec:method_render}, we present the proposed differentiable rendering of 3DGS for fisheye cameras;
finally, in \cref{sec:method_unigaussian}, we introduce the proposed UniGaussian framework for driving scene reconstruction.

\begin{figure*}[t!]
    \centering
    \begin{subfigure}[b]{0.58\textwidth}
        \includegraphics[width=\textwidth]{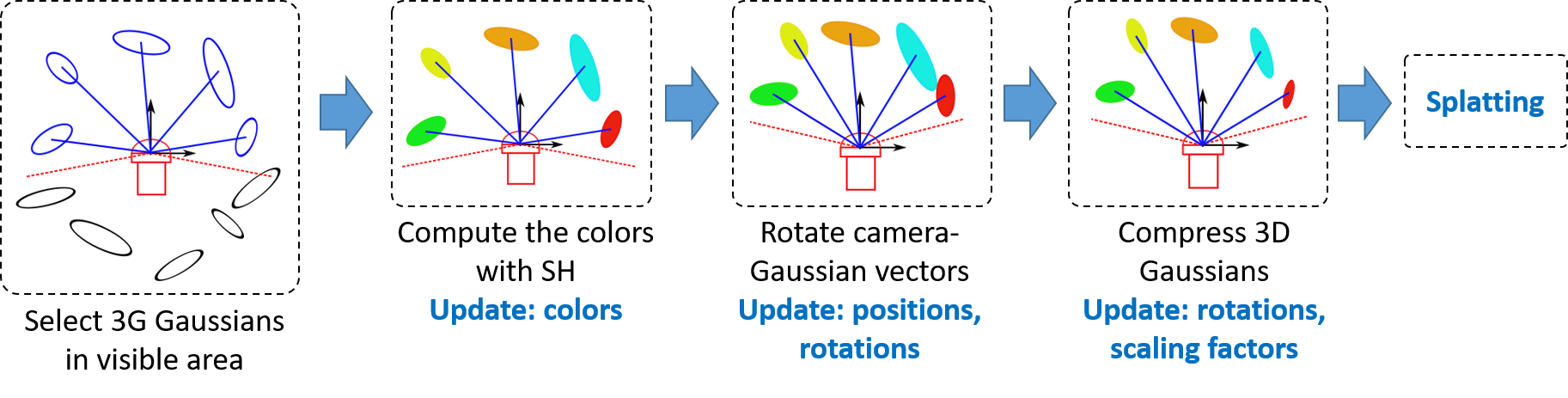}
        \caption{The architecture of our differentiable rendering for fisheye cameras.}
        \label{fig:flowchart}
    \end{subfigure}
    \begin{subfigure}[b]{0.1\textwidth}
        \includegraphics[width=\textwidth]{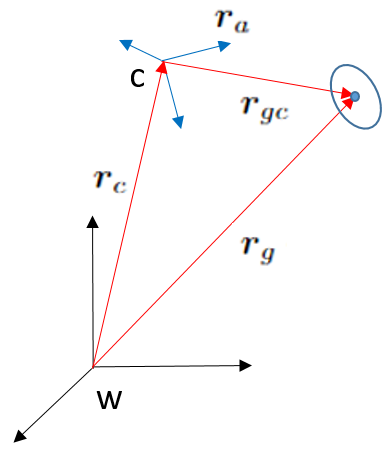}
        \caption{3D Gaussian translation \& rotation.}
        \label{fig:move_3dgs_a}
    \end{subfigure}
    \begin{subfigure}[b]{0.15\textwidth}
        \includegraphics[width=\textwidth]{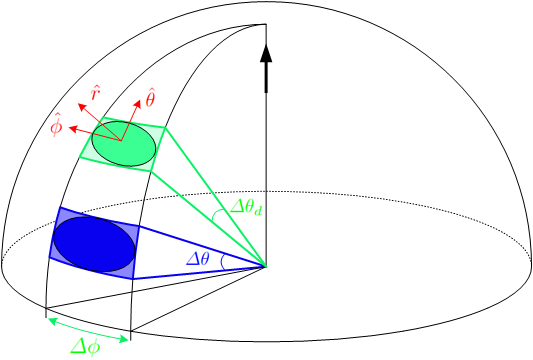}
        \caption{3D Gaussian stretching.}
        \label{fig:move_3dgs_b}
    \end{subfigure}
    \vspace{-0.3cm}
    \caption{Illustrations of our 3DGS rendering with fisheye cameras.}
    \label{fig:move_3dgs}
\end{figure*}

\subsection{Mathematical Models of Fisheye Cameras}
\label{sec:method_math}

The Kannala-Brandt~\cite{kannala2006generic} and MEI~\cite{mei2007single} mathematical models are two models commonly used for fisheye cameras.
Each model consists of two main steps.
The first step can be referred to as the ``mirror transformation'' that characterizes the twisting of light rays induced by the lens/mirrors or the transformation that translates 3D points to new positions by rotating the camera-point vectors towards the optical axis of the camera.
The second step is projecting 3D points from new positions onto the image plane.
This step utilizes either the camera intrinsic matrix or a pseudo intrinsic matrix.

Let $\theta$ and $\theta_d$ be the included angles between the camera-point vectors and the optical axis before and after applying the rotation, respectively.
Then, the first step is defined as: 
\begin{equation}
\label{eq:mirror_trans} 
  \theta_d = \mathcal{M}(\theta),     
\end{equation}
where $\mathcal{M}$ is the transformation specified by different models.
For the Kannala-Brandt model, Eq.~\eqref{eq:mirror_trans} is defined as:
\begin{small}
\begin{equation}
    % \small
    \label{eq:Kannala-Brandt}
    \begin{split}
    \theta_d = \arctan r_d 
            = \arctan \left( \theta \left(1+k_1\theta^2 + k_2\theta^4 + k_3\theta^6 + k_4\theta^8 \right) \right),
    \end{split}
\end{equation}
\end{small}%
where $k_i$ is a distortion coefficient.
For the MEI model, Eq.~\eqref{eq:mirror_trans} is defined as:
\begin{equation}
    \begin{split}
    \theta_d = \arctan r_d = \arctan \left( \chi + k_1 \chi^3 + k_2 \chi^5 \right),
    \end{split}
    \label{eq:MEI_theta_d}
\end{equation}
where $\chi = \frac{\sin{\theta}}{\cos{\theta}+\xi}$ and the twist of light rays is controlled by three parameters $k_1$, $k_2$ and $\xi$.
Next, these translated 3D points are projected onto the image plane with intrinsic parameters $f_x$, $f_y$, $u_0$ and $v_0$. 
Specifically, the pseudo intrinsic matrix defined in Eq.~\eqref{eq:intrinsic} is used to project the translated 3D points onto the image plane.
Since the focal length $f_x$, $f_y$ and the gain $\eta$ cannot be estimated independently, $\gamma_1$ and $\gamma_2$ are used as the pseudo focal length of the MEI model.
\begin{small}
\begin{equation}
  % \small
  \begin{bmatrix}
   \gamma_1 & 0 & u_0 \\
   0 & \gamma_2 & v_0 \\
   0 &        0 & 1   \\
  \end{bmatrix} =   \begin{bmatrix}
   f_x \eta & 0 & u_0 \\
   0 & f_y \eta & v_0 \\
   0 &        0 & 1   \\
  \end{bmatrix}.
  \label{eq:intrinsic}
\end{equation}
\end{small}%
Note that the MEI model also considers the tangential distortion caused by the misalignment between the camera optical axis and the mirrors/lens rotational axis~\cite{mei2007single}, but the influence of this distortion is minor so we neglect it here.

\vspace{-0.3cm}
\paragraph{Discussion.}
From Eqs.~\eqref{eq:Kannala-Brandt} and \eqref{eq:MEI_theta_d}, we can see that the principles of these two camera models are similar.
We can move 3D points by rotating the camera-point vectors towards the optical axis of the camera and then project these 3D points onto the image plane with the pinhole camera model.
From this perspective, the rendering of a scene represented by 3D Gaussians follows a similar procedure, with the main differences being that 3D points are replaced by 3D Gaussians and the rotational transformation of the camera-point vectors introduces the deformation to 3D Gaussians.
In light of this, we formulate a transformation method to approximate the deformation of 3D Gaussians for fisheye cameras while maintaining differentiability.

\subsection{3DGS Rendering with Fisheye Cameras}
\label{sec:method_render}

\paragraph{Overview.}
The architecture of the proposed differentiable rendering method is shown in \cref{fig:flowchart}.
Suppose a driving scene is represented by a set of 3D Gaussians. 
When rendering from a specific viewpoint, we first select all visible 3D Gaussians and compute their colors utilizing spherical harmonics in the world space.
Subsequently, we adjust the positions of the selected 3D Gaussians via rotating the camera-Gaussian center vectors towards the optical axis of the camera.
Also, the poses of the selected 3D Gaussian are rotated with the camera-Gaussian center vectors.
Then, the adjusted 3D Gaussians are compressed in both polar and tangential directions.
The scaling factor and the quaternion of the 3D Gaussians are updated via eigendecomposition.
Finally, the updated 3D Gaussians are projected onto the image plane to generate the final rendering.
Overall, our method is generic to various differentiable camera models and can be used for driving scene reconstruction with large FOVs.
Besides, our method does not require large amounts of additional GPU memory, so is suitable for processing millions of 3D Gaussians in driving scene reconstruction.

\vspace{-0.3cm}
\paragraph{3D Gaussian Position and Pose Adjustment.}
Let $\bm{r}_g$ and $\bm{r}_c$ be the positions of 3D Gaussians and cameras in the world coordinate space, respectively, and let $\bm{r}_a$ be the optical axis of cameras.
As shown in \cref{fig:move_3dgs_a}, when performing the transformation, the camera-Gaussian center vector needs to be rotated by $\theta_{\Delta}$ around the vector $\bm{r}_{rot}$.
Here,  $\theta_{\Delta} = \theta_d - \theta$, and $\bm{r}_{rot}$ is denoted by:
\begin{equation}
\bm{r}_{rot} = \frac{\bm{r}_{gc}}{||\bm{r}_{gc}||} \times \bm{r}_a,
\label{eq:r_rot}
\end{equation}
where $\bm{r}_{gc} = {\bm{r}_g}-{\bm{r}_c}$, $\theta$ is the included angle between the camera-Gaussian center vector and $\bm{r}_a$, and $\theta_d$ is computed using Eq.~\eqref{eq:mirror_trans}.
Then, the new position $\bm{r}'_g$ of the 3D Gaussian is computed as:
\begin{equation}
  \bm{r}_{g}' = \bm{C}(\delta q) \bm{r}_{gc} + \bm{r}_{c}, \quad \text{where}~\delta q = q_{vec}(\bm{r}_{rot}, \theta_{\Delta}),
  \label{eq:rg}
\end{equation}
and $q_{vec}$ denotes the transform from the axis-angle to the quaternion and $\bm{C}(\cdot)$ is the transform from the quaternion to the rotation matrix.
Since the pose of the 3D Gaussian needs to be rotated along the camera-Gaussian center vector, the new quaternion $q'_g$ is defined as:
\begin{equation}
  q_g' = \delta q \otimes q_g,  
\end{equation}
where $q_g$ is the original quaternion and $\otimes$ denotes the quaternion multiplication.

\vspace{-0.3cm}
\paragraph{Compression of 3D Gaussians.}
Corresponding to the compression of FOV, 3D Gaussians need to be compressed in the polar and tangential directions;
otherwise, adjacent 3D Gaussians will have greater degree of overlap on the splatting plane.
To compress 3D Gaussians in the given direction, we use a stretching matrix $\bm{S}$ which is defined as:
\begin{small}
\begin{equation}
% \small
\bm{S}(\hat{\bm{n}},k) = \begin{bmatrix} 1+(k-1)n_x^2, & (k-1)n_xn_y, & (k-1)n_xn_z \\ (k-1)n_xn_y, & 1+(k-1)n_y^2, & (k-1)n_yn_z \\ (k-1)n_xn_z, & (k-1)n_yn_z, & 1+(k-1)n_z^2 \end{bmatrix},
\end{equation}
\end{small}%
where $\hat{\bm{n}}$ is the stretching direction and $k$ is the stretching ratio.
This stretching matrix $\bm{S}$ is used to scale targets along an arbitrary axis~\cite{dunn20113d}.
In \cref{fig:move_3dgs_b}, the blue and green ellipsoids denote 3D Gaussians before and after translation, rotation, and stretching.
These two ellipsoids have the same tangential field angle $\Delta \phi$.
The tangential stretching ratio is computed by $\sin \theta_d / \sin \theta$ and the polar stretching ratio is defined as:
\begin{equation}
k_\theta = \Delta \theta_d / \Delta \theta,
\end{equation}
where $\Delta \theta = \max(||\hat{\theta}\cdot \mathbf{v}_x||, ||\hat{\theta}\cdot \mathbf{v}_y||, ||\hat{\theta}\cdot \mathbf{v}_z||) \times 2$, $\hat{\theta}$ is the local orthogonal unit vector points to the direction of optical axis,
$\mathbf{v}_x$, $\mathbf{v}_y$ and $\mathbf{v}_z$ are the principal axes of the 3D Gaussian with scale.
$\Delta \theta_d$ can be computed with the Taylor series expansion of Eq.~\eqref{eq:mirror_trans} as:
\begin{equation}
  \Delta \theta_d = \frac{d\theta_d}{d \theta} \Delta \theta + \frac{1}{2!}\frac{d^2\theta_d}{d \theta^2} \Delta \theta^2 +  \frac{1}{3!}\frac{d^3\theta_d}{d \theta^3} \Delta \theta^3 + \cdots.
\end{equation}
Therefore, for the Kannala-Brandt model, we have:
\begin{equation}
  \frac{d\theta_d}{d\theta}=\frac{1}{1+r_d^2}\frac{dr_d}{d\theta},
\end{equation}
\begin{equation}
  \frac{d^2\theta_d}{d\theta^2} = \frac{\frac{d^2r_d}{d\theta^2} (1+r_d^2)-2r_d (\frac{dr_d}{d\theta})^2}{(1+r_d^2)^2},
\end{equation}
where $\frac{dr_d}{d\theta} = 1+3k_1\theta^2+5k_2\theta^4+7k_3\theta^6+9k_4\theta^8$
and $\frac{d^2r_d}{d\theta^2} = 6k_1\theta+20k_2\theta^3+42k_3\theta^5+72k_4\theta^7$.
While for the MEI model, we have:
\begin{equation}
  \frac{d\theta_d}{d\theta}=\frac{(1+3k_1 \chi^2 + 5k_2 \chi^4)}{(1+r_d^2)}\frac{d \chi}{d\theta},
\end{equation}
\begin{equation}
  \frac{d^2\theta_d}{d\theta^2}=\frac{\mathcal{U}-\mathcal{V}}{(1+r_d^2)^2},
\end{equation}
\begin{equation}
  \mathcal{V} = 2r_d(1+3k_1 \chi + 5k_2 \chi^4)^2 (\frac{d \chi}{d \theta})^2,
\label{eq:mei_v}
\end{equation}
\begin{small}
\begin{equation}
  % \small
  \mathcal{U}=((6k_1 \chi + 20k_2 \chi^3) (\frac{d \chi}{d \theta})^2 + 
  (1{+}3k_1 \chi^2{ + }5k_2 \chi^4) \frac{d^2 \chi}{d \theta^2})(1{+}r_d^2),
  \label{eq:mei_w}
\end{equation}
\end{small}%
where $\frac{d \chi}{d\theta} = \frac{1+ \xi \cos \theta}{(\cos \theta + \xi)^2}$ and $\frac{d^2 \chi}{d\theta^2} = \frac{\sin \theta (2 + \xi \cos \theta - \xi^2)}{(\cos \theta + \xi)^3}$.
Here, $d^n\theta_d/d \theta^n (n>2)$ also have analytical expressions for both the Kannala-Brandt and MEI models.
When the linear approximation is used, $k_\theta = d \theta_d/d \theta$ so the computation is considerably simplified.

\vspace{-0.3cm}
\paragraph{Update Scaling Factors and Rotations.}
Generally, random variables that conform to a Gaussian distribution can retain this property even after applying affine transformations.
Since the stretching operation is an affine transformation in our method, the covariance matrix of the resultant Gaussian distribution is defined as:
\begin{equation}
  \bm{\Sigma}'=\bm{S}_{\theta}\bm{S}_{\phi}\bm{\Sigma}\bm{S}_{\phi}^T\bm{S}_{\theta}^T,
\end{equation}
where $\bm{S}_{\theta}$ and $\bm{S}_{\phi}$ are the stretching matrices in the polar and tangential directions.
Besides, the three scaling factors and the quaternion of the new 3D Gaussian are defined as:
\begin{equation}
  s_i = \sqrt{\lambda_i}, \quad i{\in}\{1,2,3\},
\end{equation}
\begin{equation}
  \label{eq:eigen}
  \bm{q}=quat\left( \left[ \bm{v_1}, \bm{v_2}, \bm{v_1}\times\bm{v_2} \right] \right),
\end{equation}
where $\lambda_i$ is the eigenvalue of $\Sigma'$, $\bm{v}_i$ is the eigenvector of $\Sigma'$, and $\bm{v_3}$ is replaced by $\bm{v_1}\times\bm{v_2}$ to ensure that these vectors form a right-hand system.

\subsection{UniGaussian for Driving Scene Reconstruction}
\label{sec:method_unigaussian}

\cref{fig:framework} depicts an overview of our approach to learning unified Gaussian representations from multiple camera models for driving scene reconstruction.

\vspace{-0.3cm}
\paragraph{Driving Scene Gaussians.}
With input images from different cameras, we represent a driving scene with composite 3D Gaussians\footnote{
In this work, ``unified Gaussian representation'' refers to learning a unified model for multiple sensors, while ``composite Gaussians'' refers to scene decomposition in the same model.
}.
Following~\cite{zhou2024hugs,zhou2024drivinggaussian,yan2024street}, we decompose the scene into background Gaussians for static scenes, dynamic Gaussians for moving objects and sky Gaussians for distant regions.
Besides, we employ LiDAR point clouds to initialize 3D Gaussians by accumulating all LiDAR frames and projecting points onto images for color extraction.
This provides a better representation of the scene geometry and enables the optional LiDAR simulation in our framework.

\vspace{-0.3cm}
\paragraph{Modeling Multiple Sensors and Modalities.}
To account for the distortion of fisheye images, we applied a series of affine transformations to further process 3D Gaussians for fisheye cameras as introduced in \cref{sec:method_render} while pinhole cameras skip these transformations.
These processed 3D Gaussians are rendered with a unified tile-based rasterizer.
Then, to resolve the exposure difference between pinhole and fisheye cameras, we applied camera-dependent scaling and biases factors to the rendering images as~\cite{zhou2024hugs}.
This models appearance difference between pinhole and fisheye cameras.
Besides, we also render other modalities, including depth maps, semantic maps, and normal maps.
To render these maps, each 3D Gaussian is added with the corresponding 3D logits, and then 2D maps are obtained via $\alpha$-blending of these 3D logits in the rasterizer.
Although LiDAR simulation may not be directly achieved by explicit 3D Gaussians, we can obtain point clouds from the rendering depth maps by extracting points based on real-world LiDAR parameters and scans.
Similarly, the intensities of point clouds can be obtained by generating intensity maps from the rasterizer and extracting the corresponding points.
Please refer to the supplementary material for more details of the optional LiDAR simulation.

\vspace{-0.3cm}
\paragraph{Adaptive Density Control.}
Driving scene reconstruction usually requires millions of 3D Gaussians.
To enhance the adaptive density control of these Gaussians, we employ the Markov Chain Monte Carlo (MCMC) sampling and relocation strategy following~\cite{kheradmand20243d}.
Specifically, we consider 3D Gaussian densification and pruning as a deterministic state transition of MCMC samples and employ a relocation strategy to dynamically adjust the position of Gaussians while preserving sample probability.
With the opacity and scaling regularization, this adaptive control encourages the removal of redundant Gaussians and produces more compact scene Gaussians.
Please refer to~\cite{kheradmand20243d} for more details.

\vspace{-0.3cm}
\paragraph{Model Optimization.}
We employ multiple losses to optimize our model.
These losses constrain the shared 3D Gaussian representation to learn complementary information from different modalities, ensuring the consistency and complementarity of outputs across sensors and modalities.
Overall, the training loss $\mathcal{L}$ is defined as:
\begin{equation}
    \mathcal{L} = \mathcal{L}_{rgb}^{P} + \mathcal{L}_{rgb}^{F} + \mathcal{L}_{d} \\
    + \mathcal{L}_{s} + \mathcal{L}_{n} + \mathcal{L}_{reg},
\label{eq:loss}
\end{equation}
where $\mathcal{L}_{rgb}^{P}$ and $\mathcal{L}_{rgb}^{F}$ are the reconstruction losses between the ground-truth and the rendering pinhole/fisheye images following~\cite{kerbl3Dgaussians},
$\mathcal{L}_{d}$ is the depth loss computed between the rendering depth and the monocular depth~\cite{hu2024metric3d} or LiDAR depth,
$\mathcal{L}_{s}$ is the semantic loss computed with the rendering semantic map and the predefined 2D semantic segmentation map~\cite{Liao2022PAMI},
$\mathcal{L}_{n}$ is the normal consistency loss regularizing the rendering normal and the normal derived from the depth,
and $\mathcal{L}_{reg}$ is the Gaussian opacity and scale regularization term~\cite{kheradmand20243d} to encourage a compact Gaussian representation.
Please refer to the supplementary material for more details.

\begin{figure}[t]
  \centering
  \resizebox{0.47\textwidth}{!}{    
   \includegraphics[width=0.8\textwidth]{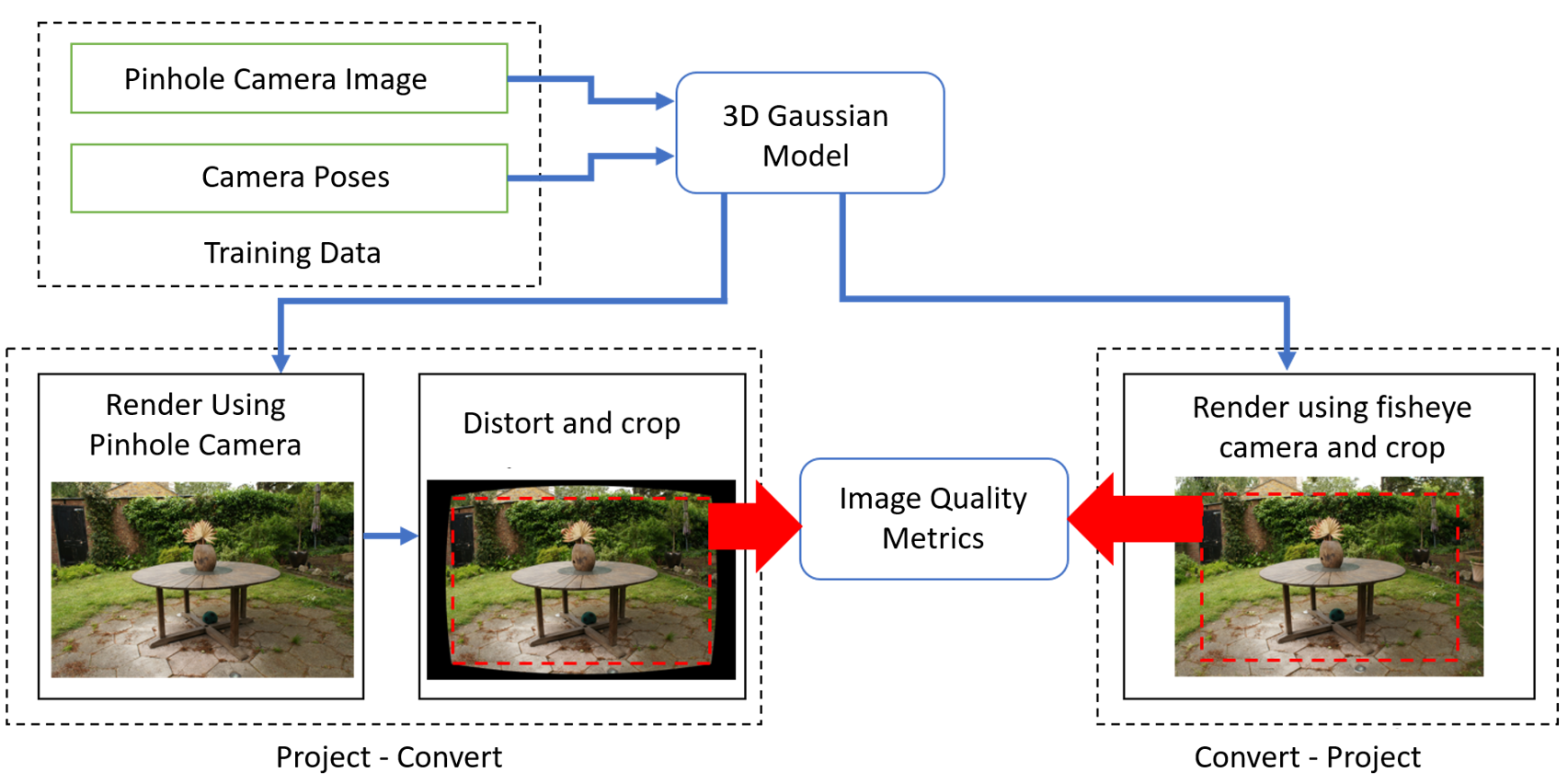}     
  }
  \vspace{-0.3cm}
  \caption{The flowchart for analyzing geometric errors.}
   \label{fig:exp_error_analysis}
 \end{figure}

\begin{table}[t!]
  \centering
  \resizebox{0.47\textwidth}{!}{
  \begin{tabular}{cc|cccc|cccc}
    \hline
        & & \multicolumn{4}{c|}{BICYCLE $FOV=56^\circ$} & \multicolumn{4}{c}{GARDEN $FOV=68^\circ$ } \\ 
        & & \multicolumn{4}{c|}{$(W,H){=}(4946,3286)~ROI(4391,2917)$} & \multicolumn{4}{c}{$(W,H){=}(5187,3361)~ROI(4358,2824)$} \\
    \hline
    $\hat{\phi}$ & $\hat{\theta}$ & PSNR$\uparrow$ & SSIM$\uparrow$ & LPIPS$\downarrow$ & Time(ms) 
    & PSNR$\uparrow$ & SSIM$\uparrow$ & LPIPS$\downarrow$ & Time(ms) \\
    \hline
    \XSolidBrush & \XSolidBrush & 29.340 & 0.926 & 0.0386 & 122 & 29.805 & 0.902 & 0.0523 & 119 \\
    \XSolidBrush & 1 & 30.752 & 0.936 & 0.0325 & 132 & 30.148 & 0.911 & 0.0455 & 124 \\
    1 & \XSolidBrush & 29.440 & 0.926 & 0.0381 & 128 & 29.835 & 0.903 & 0.0522 & 122  \\
    1 & 1 & 30.794 & 0.936 & 0.0323 & 149 & 30.189 & 0.911 & 0.0454 & 140 \\
    1 & 2 & 30.825 & 0.936 & 0.0322 & 251 & 30.192 & 0.912 & 0.0454 & 446 \\
    \hline
  \end{tabular}
  }
  \vspace{-0.3cm}
  \caption{Rendering geometric error analysis.
  $\hat{\phi}$ and $\hat{\theta}$ denote the approximation order of tangential and polar stretching ratios.
  ``\XSolidBrush'' means the 3D Gaussian is not stretched in that direction. 
  ``1'' denotes the 1st order while ``2'' denotes the 2nd order.}
  \label{tab:psnr_error}
\end{table}

\begin{figure*}[t]
  \begin{minipage}[b]{0.18\textwidth}
    \centering
    \includegraphics[width=1.0\textwidth]{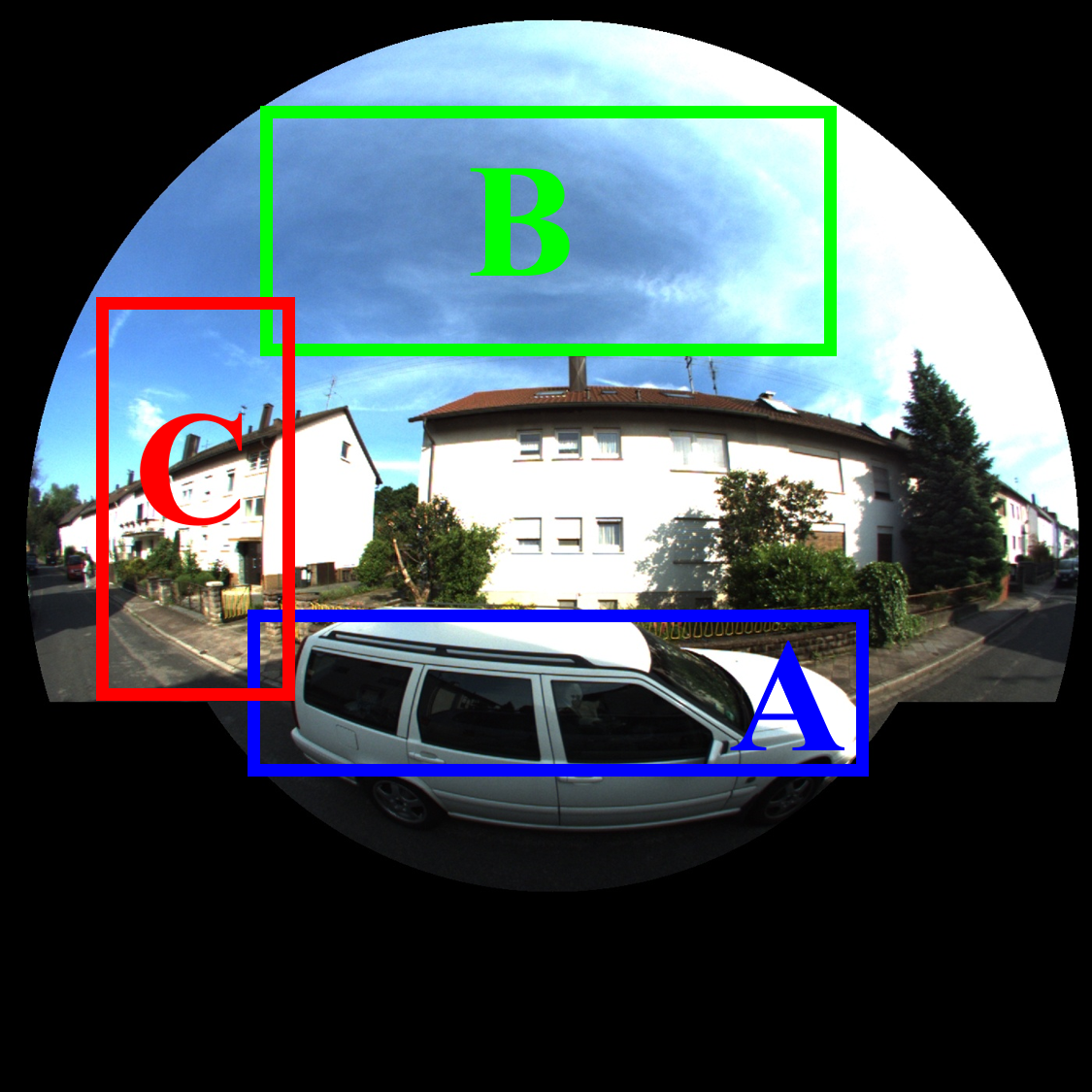} 
    \vspace{-0.3cm}
  \caption{Image zones.}
  \label{fig:small_zone}
  \end{minipage}
  \begin{minipage}[b]{0.78\textwidth}
    \centering
    \resizebox{1.0\textwidth}{!}{
    \begin{tabular}{c|c|ccccccc}
    \hline
      \multicolumn{2}{c}{} & Instant-NGP~\cite{muller2022instant} & NeRFacto-big~\cite{nerfstudio} & Zip-NeRF~\cite{barron2023zip} & 3DGS~\cite{kerbl3Dgaussians}+Undistort &  Ours (w/o-s) & Ours (w-s) \\
    \hline
     \multirow{3}{*}{ } & PSNR$\uparrow$  & 21.247 & 13.751 & 12.604 & 12.635 & 24.574 & $\underline{\bm{24.651}}$ \\
     ~ & SSIM$\uparrow$ & 0.773 & 0.548 & 0.504 & 0.508 & 0.815 & $\underline{\bm{0.817}}$ \\
     ~ & LPIPS$\downarrow$ & 0.195 & 0.306 &0.469  & 0.361 & 0.133 & $\underline{\bm{0.130}}$ \\
    \hline
     \multirow{3}{*}{Zone-A} & PSNR$\uparrow$ & 22.339 & 15.495 &15.366 & 17.010 & 22.784 & $\underline{\bm{22.919}}$  \\
     ~ & SSIM$\uparrow$  & 0.617 & 0.412 &0.405 & 0.425 & 0.688 & $\underline{\bm{0.695}}$ \\
     ~ & LPIPS$\downarrow$ & 0.371 & 0.489 &0.540 & 0.353 & 0.276 & $\underline{\bm{0.265}}$ \\
    \hline
     \multirow{3}{*}{Zone-B} & PSNR$\uparrow$  & 21.210 & 20.917 &19.228 & 12.437 & 30.830 & $\underline{\bm{30.860}}$ \\
     ~ & SSIM$\uparrow$  & 0.867 & 0.830 &0.797 & 0.710 & 0.869 & $\underline{\bm{0.869}}$  \\
     ~ & LPIPS$\downarrow$ & 0.127 & 0.155 &0.190 & 0.340 & 0.116 & $\underline{\bm{0.115}}$  \\
    \hline
     \multirow{3}{*}{Zone-C} & PSNR$\uparrow$  & 22.401 & 14.768 &11.978 & 11.637 & 25.720 & $\underline{\bm{25.861}}$  \\
     ~ & SSIM$\uparrow$  & 0.790 & 0.508 &0.426 & 0.426 & 0.838 & $\underline{\bm{0.842}}$  \\
     ~ & LPIPS$\downarrow$ & 0.154 & 0.266 &0.510 & 0.468 & 0.099 & $\underline{\bm{0.093}}$  \\
    \hline
     & Training time & $\underline{\bm{12}\text{m}}$ & 3h28m &37m & 26m & 36m & 58m  \\
     & Rendering FPS & 0.26 & 0.23 &0.28 & 121.37 & $\underline{\bm{138.89}}$ & 39.06  \\
     & 3DG Num & -- & -- & -- & 697K & 897K  & $\underline{\bm{634}\text{K}}$ \\
    \hline
    \end{tabular}
    }
    \vspace{-0.3cm}
    \captionof{table}{Results of fisheye camera simulation on KITTI-360.
    ``Ours(w-s)'' and ``Ours(w/o-s)'' denote our method w/ and w/o stretching.
    ``Undistort'' denotes fisheye images are rectangularized for reconstruction and the scene is rendered with the pinhole camera model and distorted to fisheye images.
    For a fair comparison, all methods do not use LiDAR or multimodal optimization.
    }
    \label{tab:kitti360_res}
  \end{minipage}
  \end{figure*}

  \begin{figure}[!t] 
   \centering
   \resizebox{0.45\textwidth}{!}{
   \begin{tabular}{cccc}
   \includegraphics[width=2.35cm]{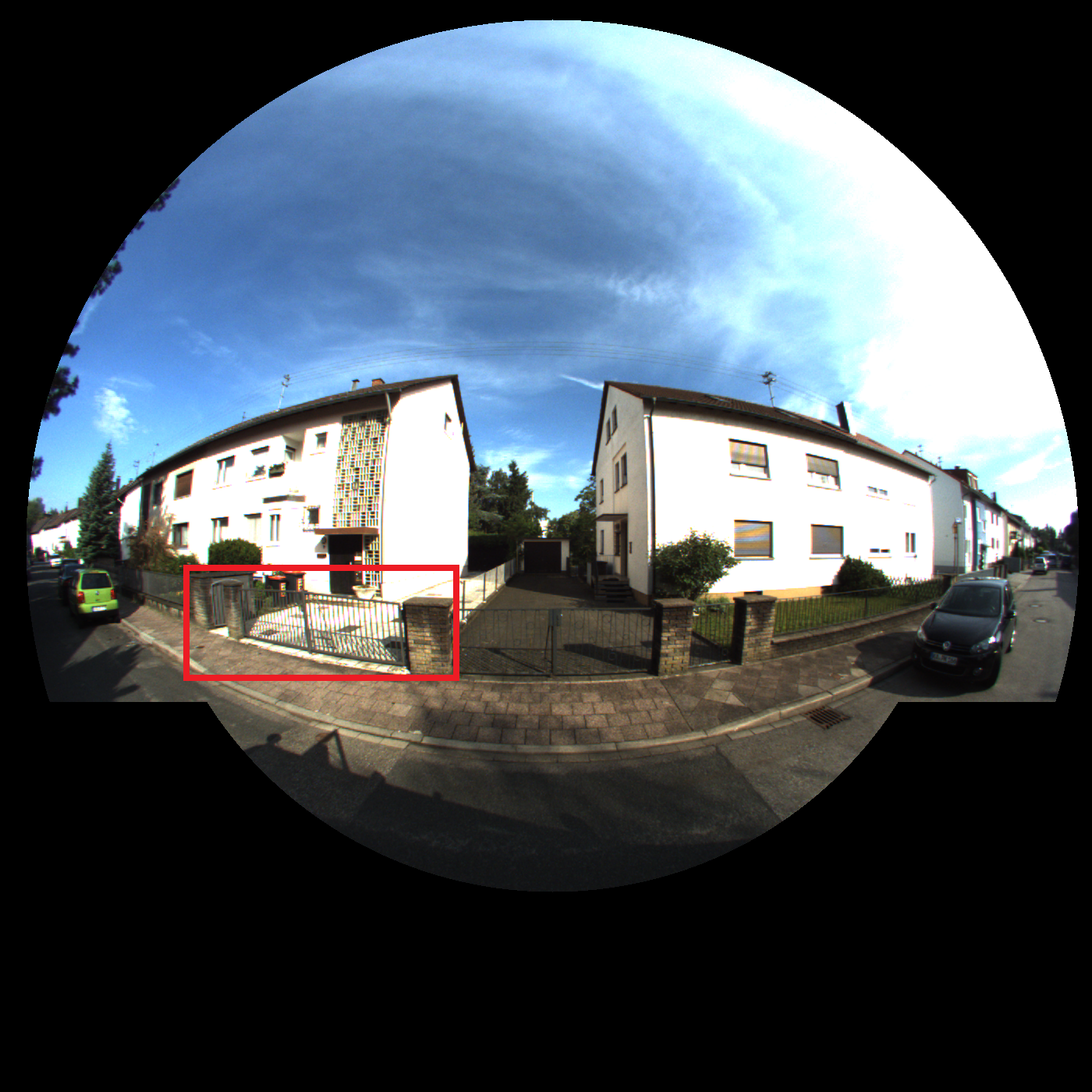}
   & \includegraphics[width=2.35cm]{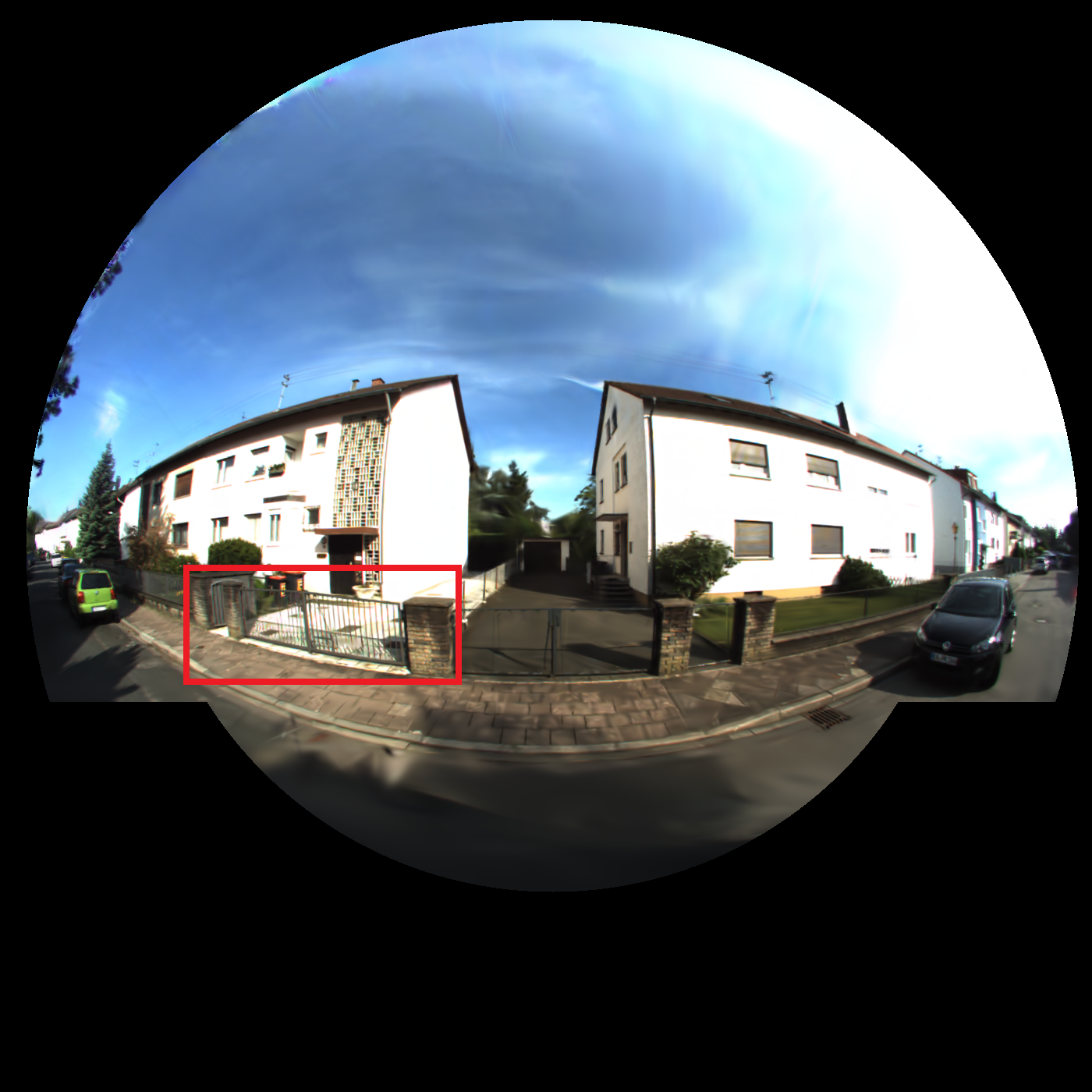}
   & \includegraphics[width=2.35cm]{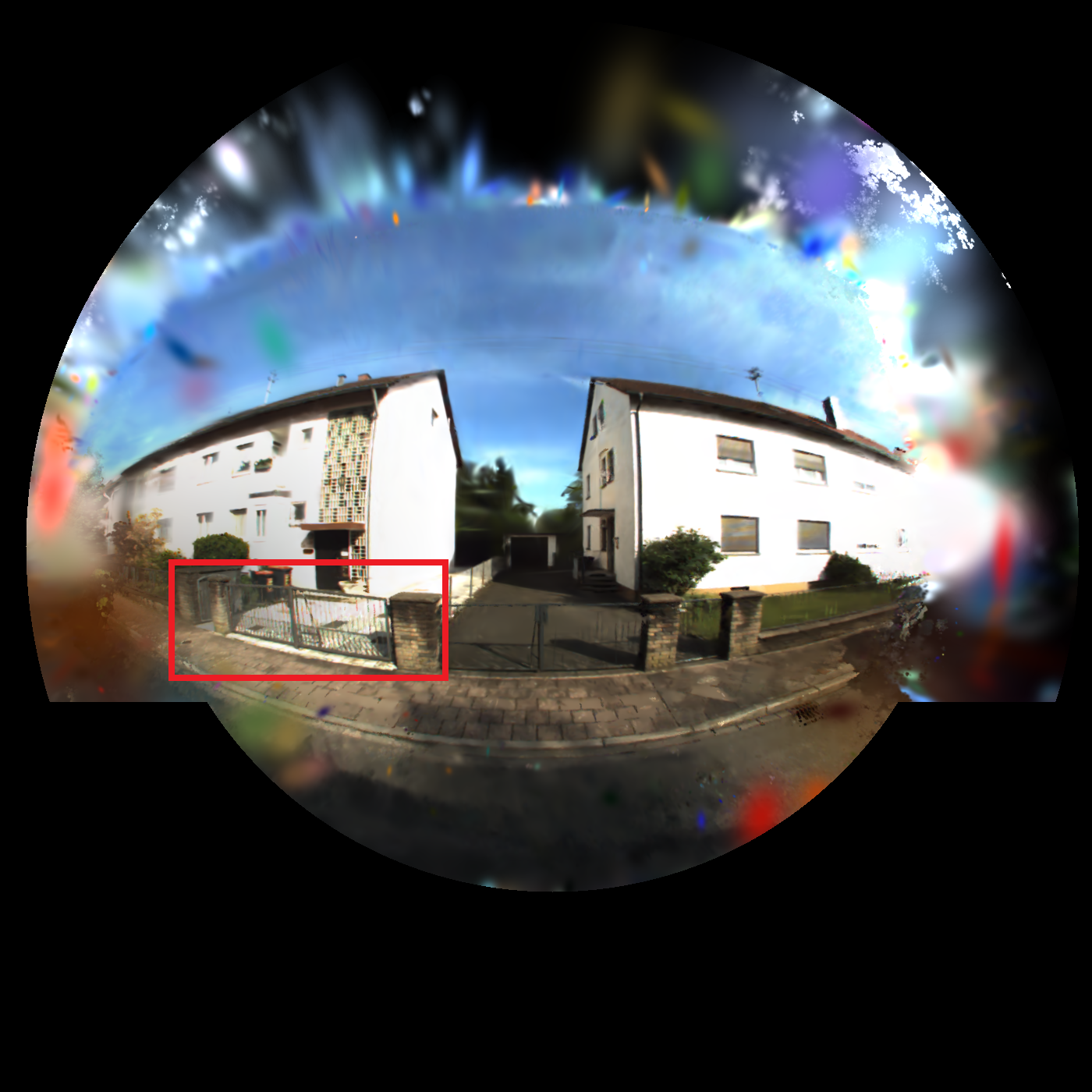}
   & \includegraphics[width=2.35cm]{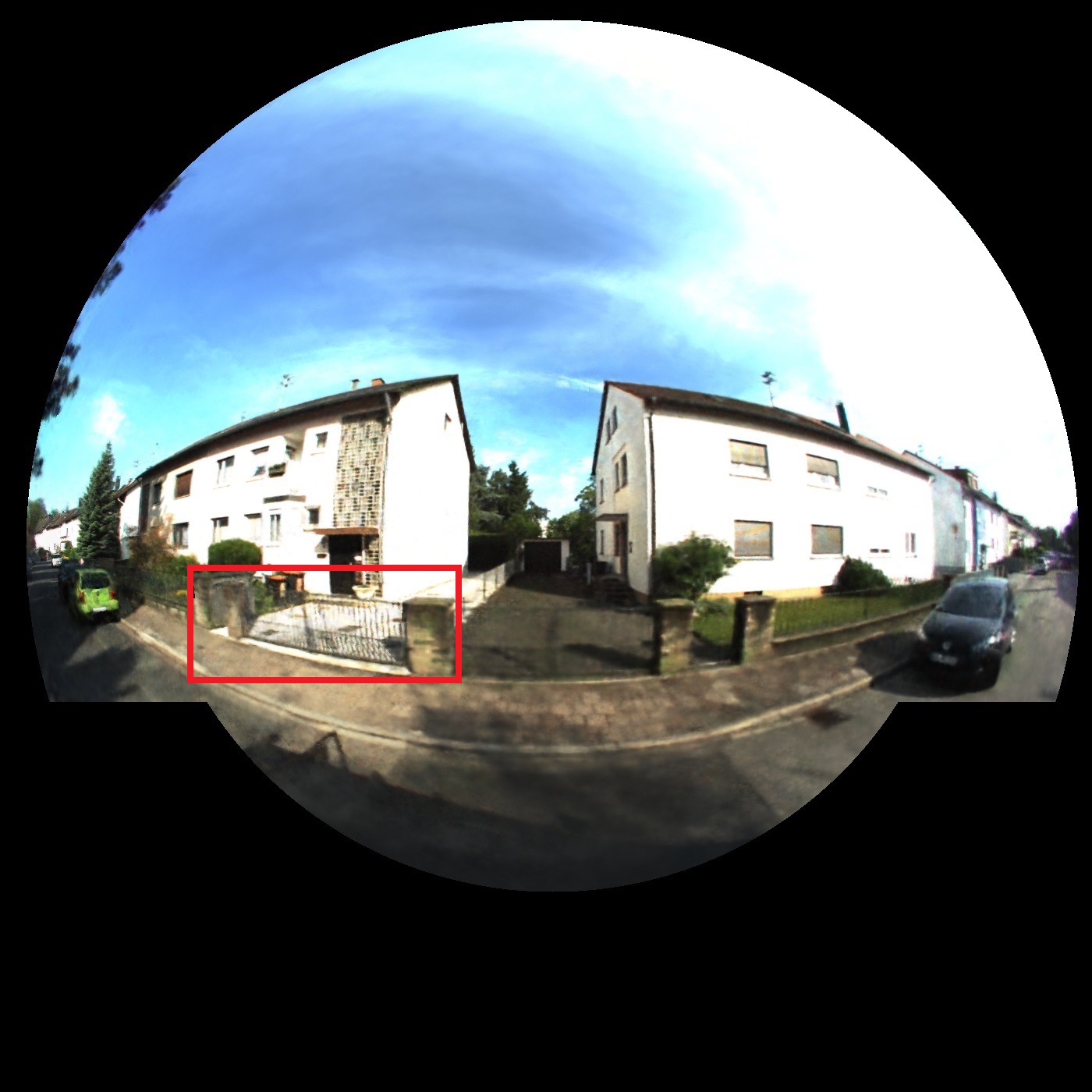} \\
  
   \includegraphics[width=2.35cm, height=1.05cm]{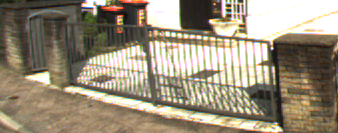}
   & \includegraphics[width=2.35cm, height=1.05cm]{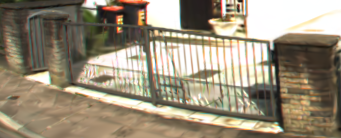}
   & \includegraphics[width=2.35cm, height=1.05cm]{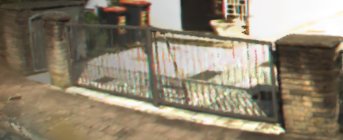}
   & \includegraphics[width=2.35cm, height=1.05cm]{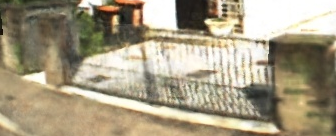} \\
  
   \includegraphics[width=2.35cm]{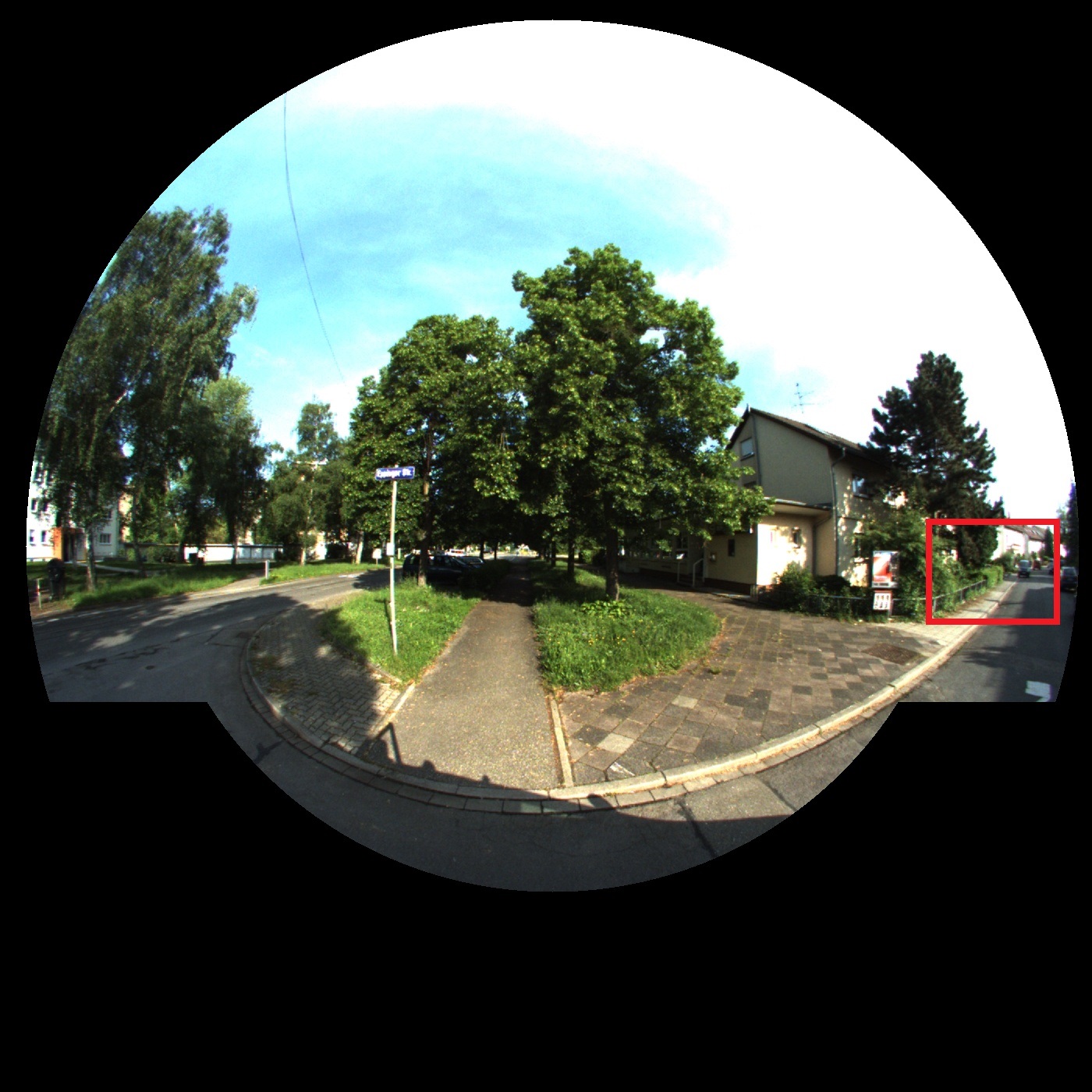}
   & \includegraphics[width=2.35cm]{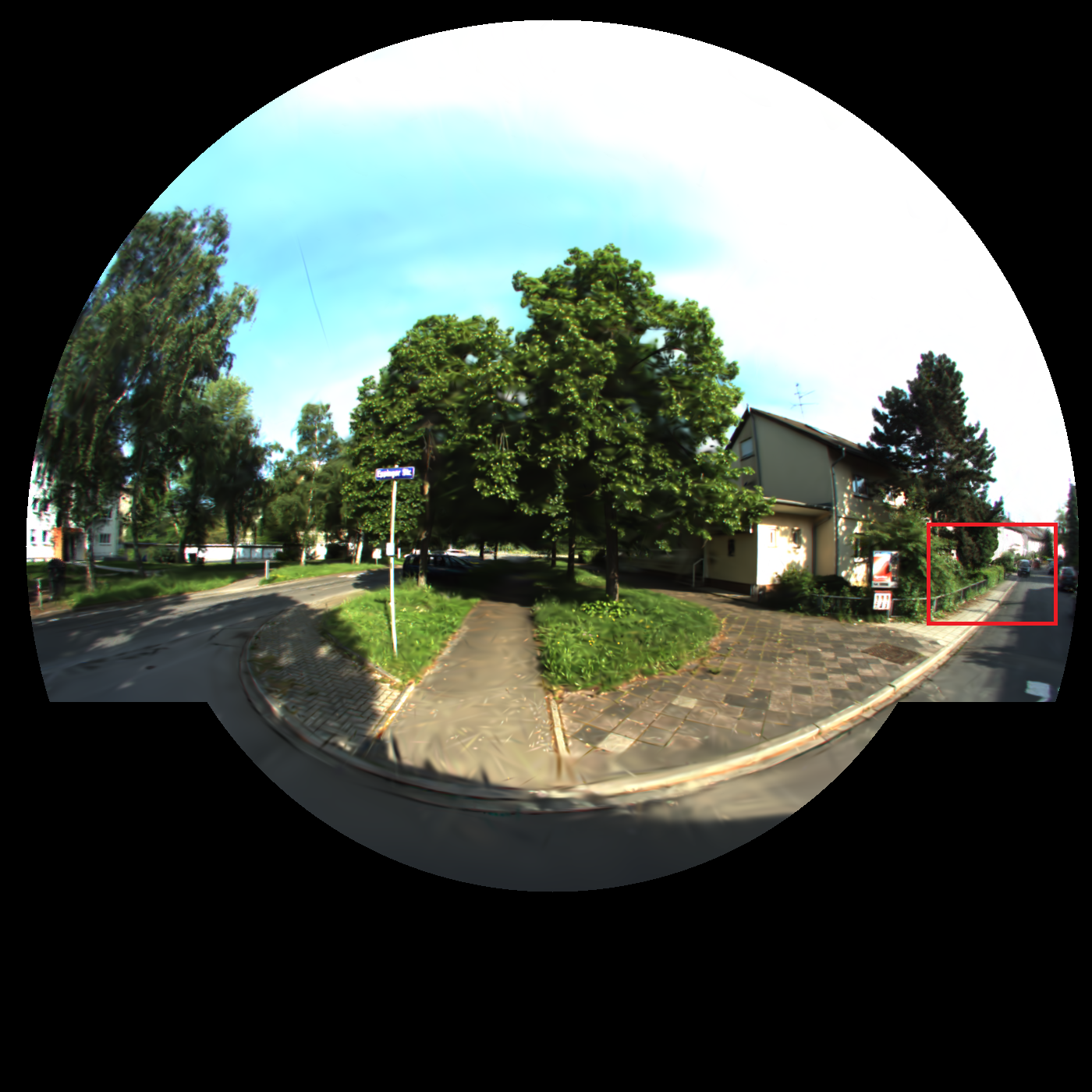}
   & \includegraphics[width=2.35cm]{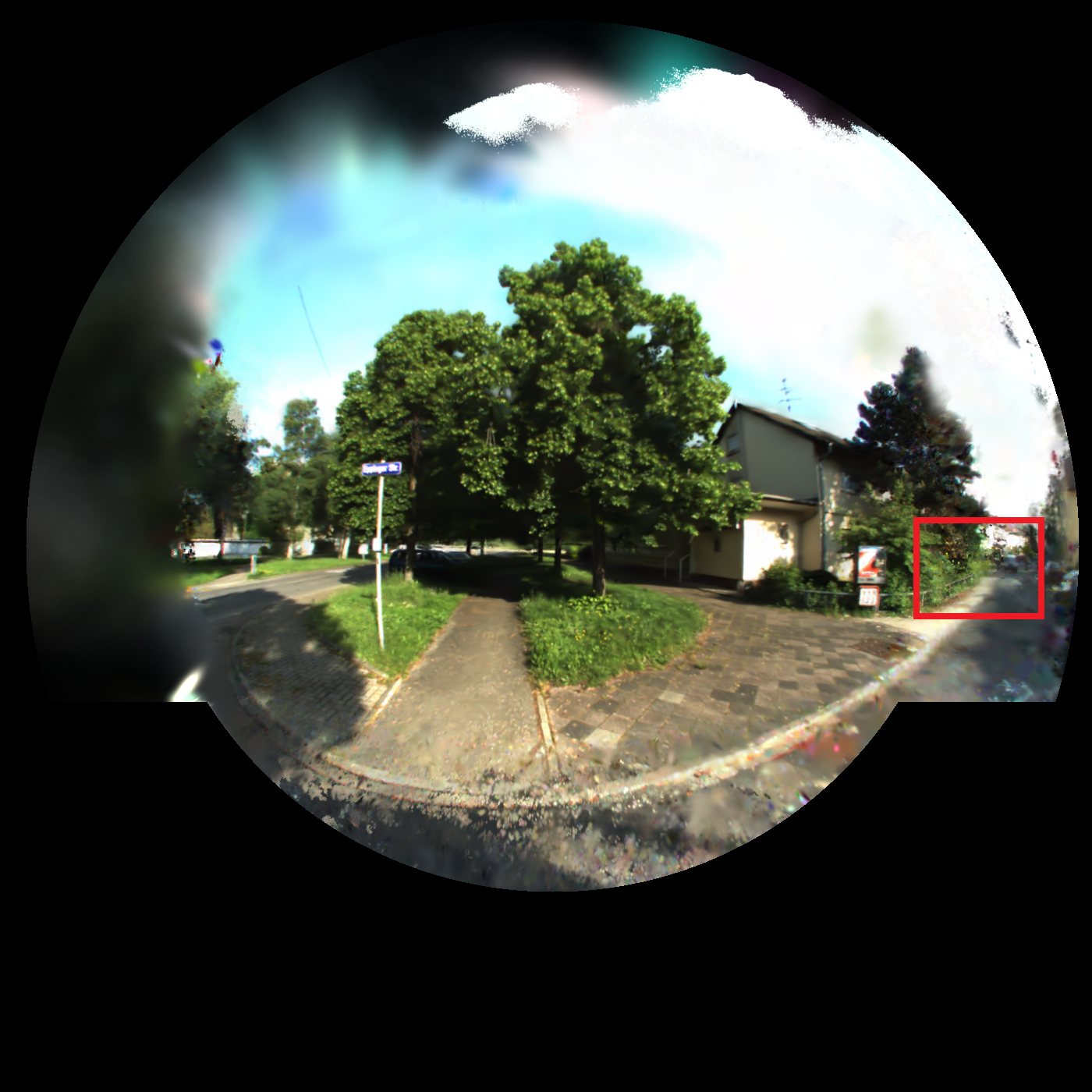}
   & \includegraphics[width=2.35cm]{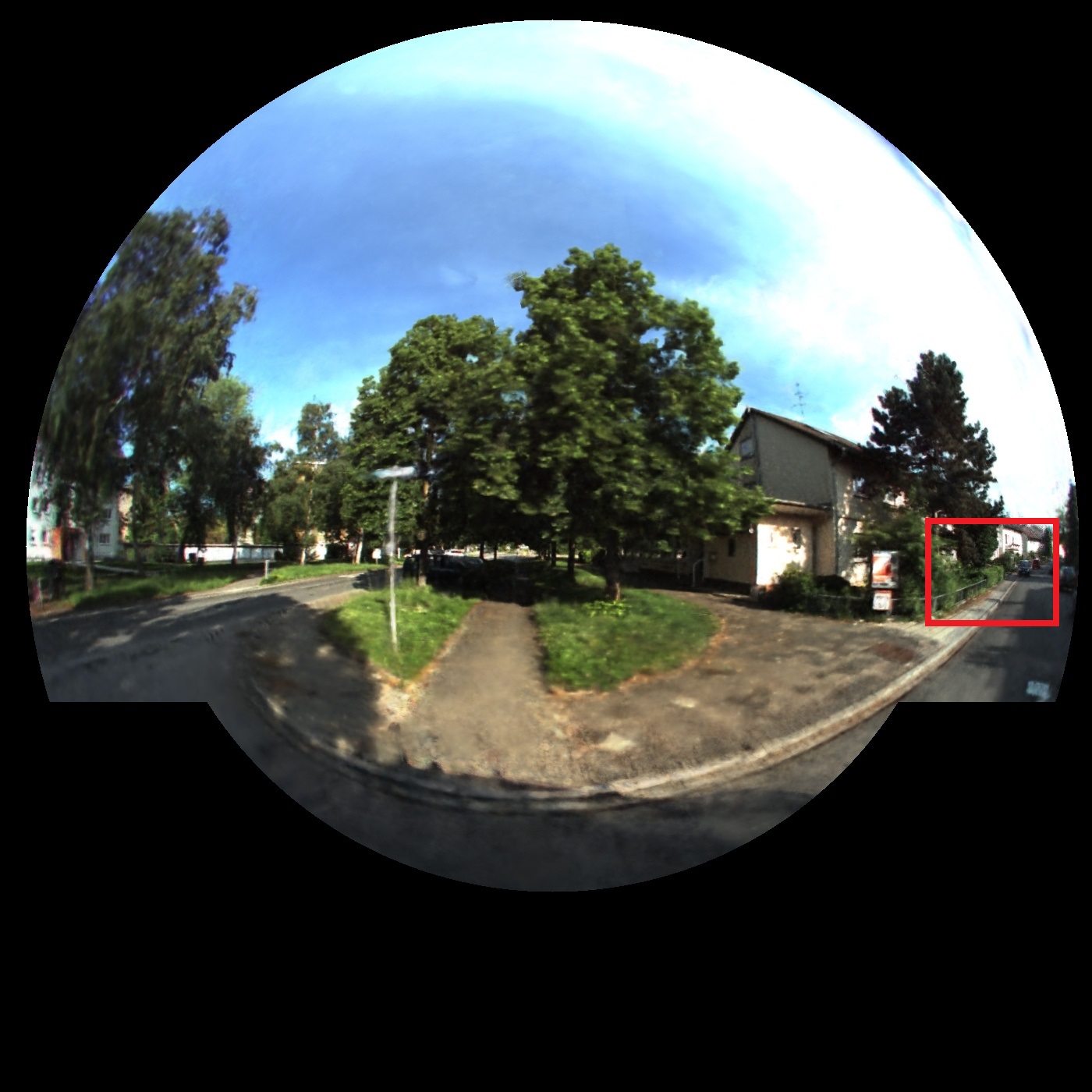} \\
  
   \includegraphics[width=2.35cm, height=1.05cm]{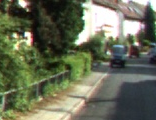}
   & \includegraphics[width=2.35cm, height=1.05cm]{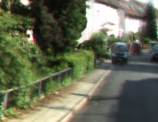}
   & \includegraphics[width=2.35cm, height=1.05cm]{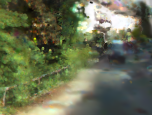}
   & \includegraphics[width=2.35cm, height=1.05cm]{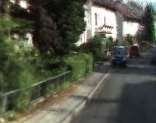} \\
  
   Ground truth & Ours & 3DGS+Undistort & Instant-NGP  \\
   \end{tabular}
   }
   \vspace{-0.3cm}
   \caption{Rendering results of fisheye simulation on KITTI-360.}
   \label{fig:kitti_render_results}
  \end{figure}

\section{Experiments}
\label{sec:experiment}

In our experiments, we first verify the effectiveness of the proposed rendering method in \cref{sec:exp_fisheye_error} and \cref{sec:exp_fisheye_simulation} and then examine the efficacy of the unified framework in \cref{sec:exp_multi_camera_simulation}.

\subsection{Fisheye Rendering Geometric Error Analysis}
\label{sec:exp_fisheye_error}

\paragraph{Relative Error Analysis.}
Since the proposed method is a ``convert-project'' method, the conversion part contains some approximations.
After the light rays are twisted by fisheye cameras, the distorted 3D Gaussian no longer follows an exact normal distribution.
The tangential stretching ration $k_\phi$ does not contain any approximation, while $k_\theta$ includes the truncation error.
When using  the first-order approximation, the truncation error of $k_{\theta}$ is defined as:
\begin{small}
\begin{equation}
  \label{eq:geometric_error}
\epsilon_{k_\theta}(\theta, \Delta \theta)=\frac{1}{2!}\frac{d^2\theta_d}{d\theta^2}\Delta \theta + \frac{1}{3!}\frac{d^3\theta_d}{d\theta^3}\Delta \theta^2+\ldots.
\end{equation}
\end{small}%
On the image plane, this can cause scale errors in the radial direction from the center of the image, so the relative error is defined as $\epsilon_{k_\theta}(\theta, \Delta \theta) \cos{\theta_d}$.
However, our design is not significantly affected by this approximation.
To verify this, we conduct the following experiment.

\vspace{-0.3cm}
\paragraph{Flowchart for Analysis.}
We employ the method depicted in \cref{fig:exp_error_analysis} to analyze the rendering geometric error introduced by these approximations.
First, a 3D Gaussian model is trained using the images of a pinhole camera.
Then, images of a fisheye camera are rendered in two different ways: the ``convert-project'' method as introduced above and the ``project-convert'' method that renders a pinhole image and distorts it to a fisheye image.
The ``project-convert'' method does not have any approximation in the conversion part, so its rendering result is considered as the reference image.
We use the ``BICYCLE'' and ``GARDEN'' of the Mip-NeRF360 dataset~\cite{barron2022mip} in this experiment because these data have enough multi-view images and accurate camera poses that enable rendering from any point of view.

\vspace{-0.3cm}
\paragraph{Results Analysis.}
The results are presented in \cref{tab:psnr_error}.
From these results, we can see that the stretching of 3D Gaussians is effective for improving the quality of rendering.
This proves the correctness of our rendering algorithm's derivation because the approximations do not degrade the image quality.
Moreover, in our method, most of the rendering time is spent on affine transformations of 3D Gaussians, while rasterization only takes around 3 ms.
Note that the time consumption shown in \cref{tab:psnr_error} is for images with large resolutions, but for autonomous driving scenes, 3D Gaussians are projected within a frustum and the image resolution is less than $4358\times2824$ so the rendering can be done in real time.
Besides, although the 2nd order approximation of the polar stretching ratio gives an improvement on rendering quality, it greatly increases the rendering time.

\begin{figure}[!tb] 
  \centering
  \resizebox{0.47\textwidth}{!}{
  \begin{tabular}{cccc}
    \includegraphics[width=2.35cm]{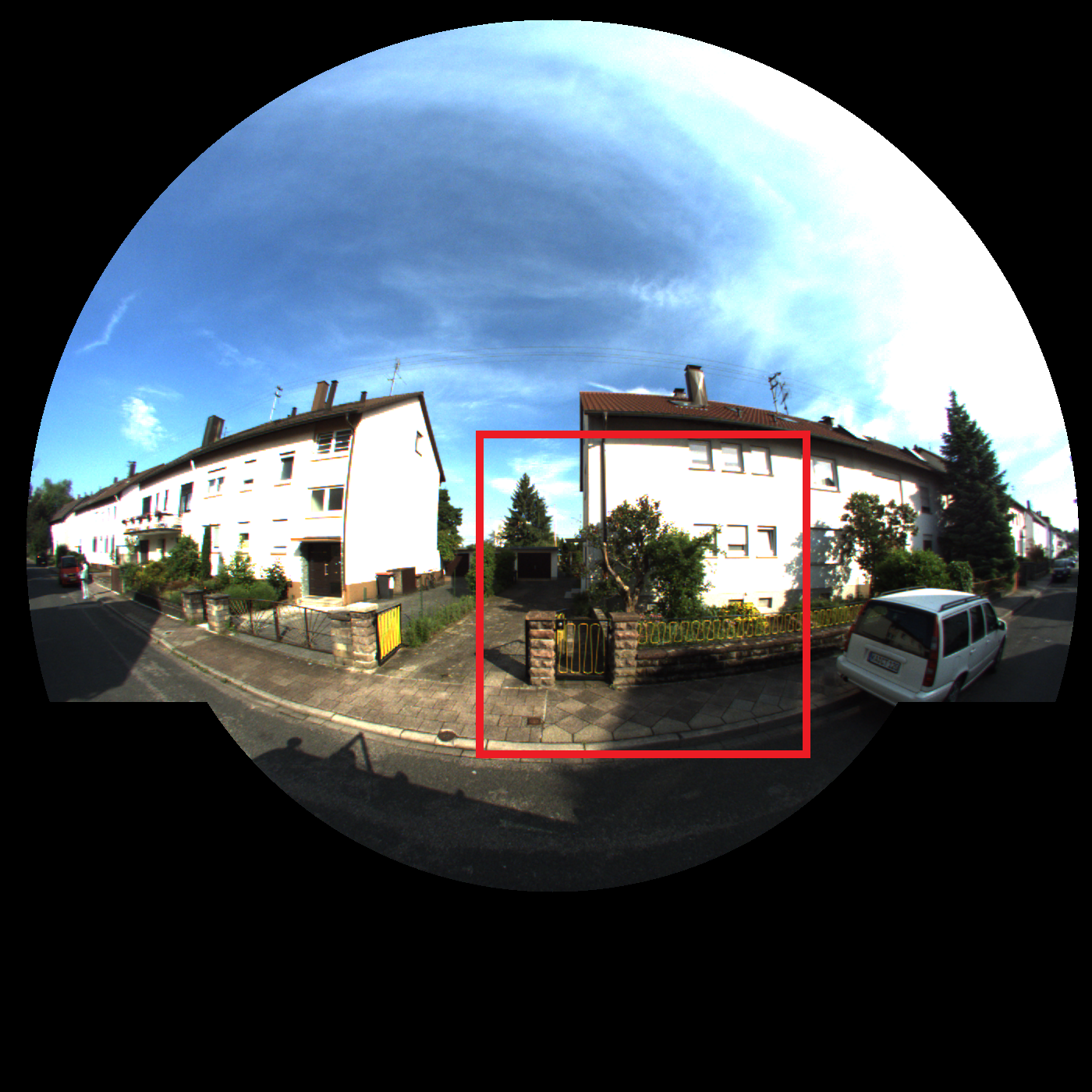}
    & \includegraphics[width=2.35cm]{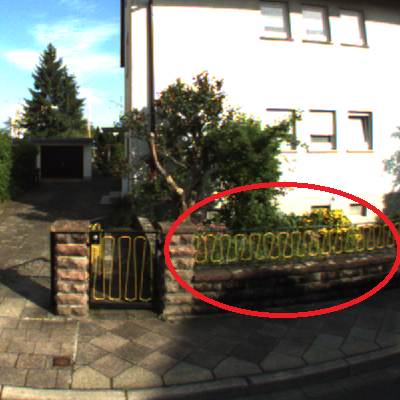}
    & \includegraphics[width=2.35cm]{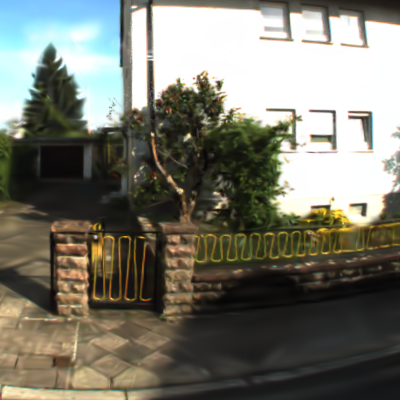}
    & \includegraphics[width=2.35cm]{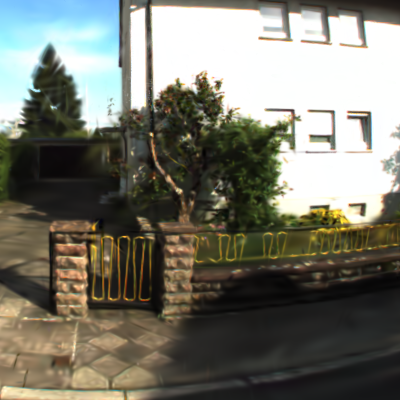} \\ 
 
    \includegraphics[width=2.35cm]{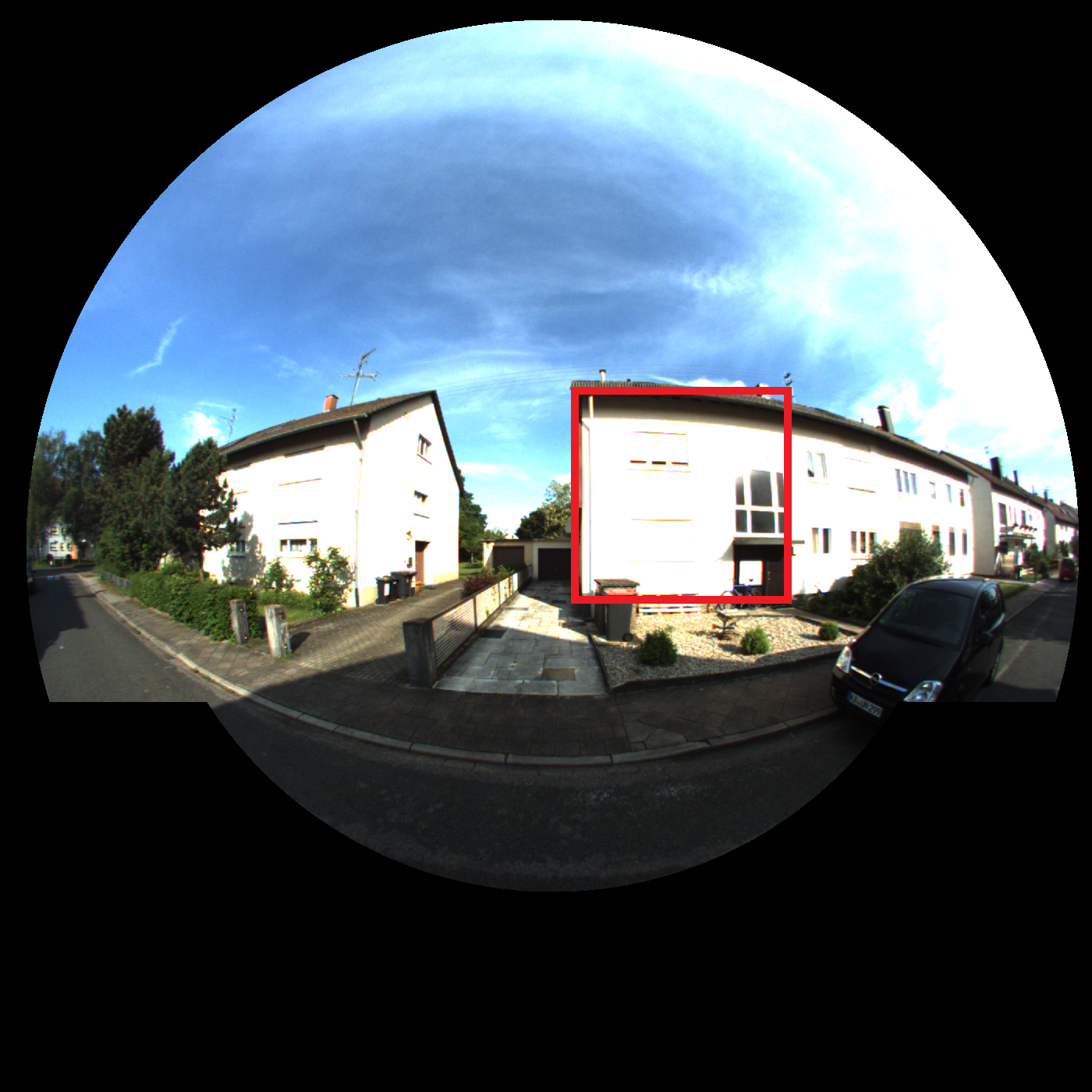}
    & \includegraphics[width=2.35cm]{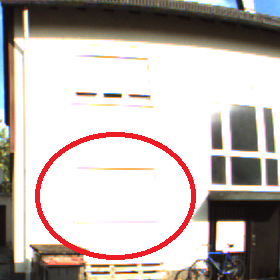}
    & \includegraphics[width=2.35cm]{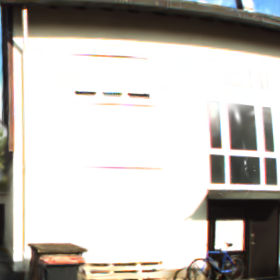}
    & \includegraphics[width=2.35cm]{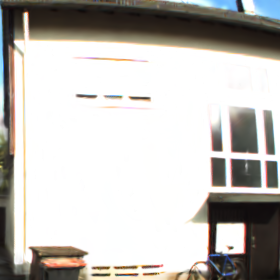} \\ 
 
   Ground truth & GT(local) & Ours(w-s) & Ours(w/o-s) \\
  \end{tabular}
  }
  \vspace{-0.3cm}
  \caption{Rendering results of our approach w/ and w/o stretching.}
  \label{fig:w_wo_stretching}
\end{figure}

\begin{figure*}[!tb] 
  \centering
  \includegraphics[width=0.85\textwidth]{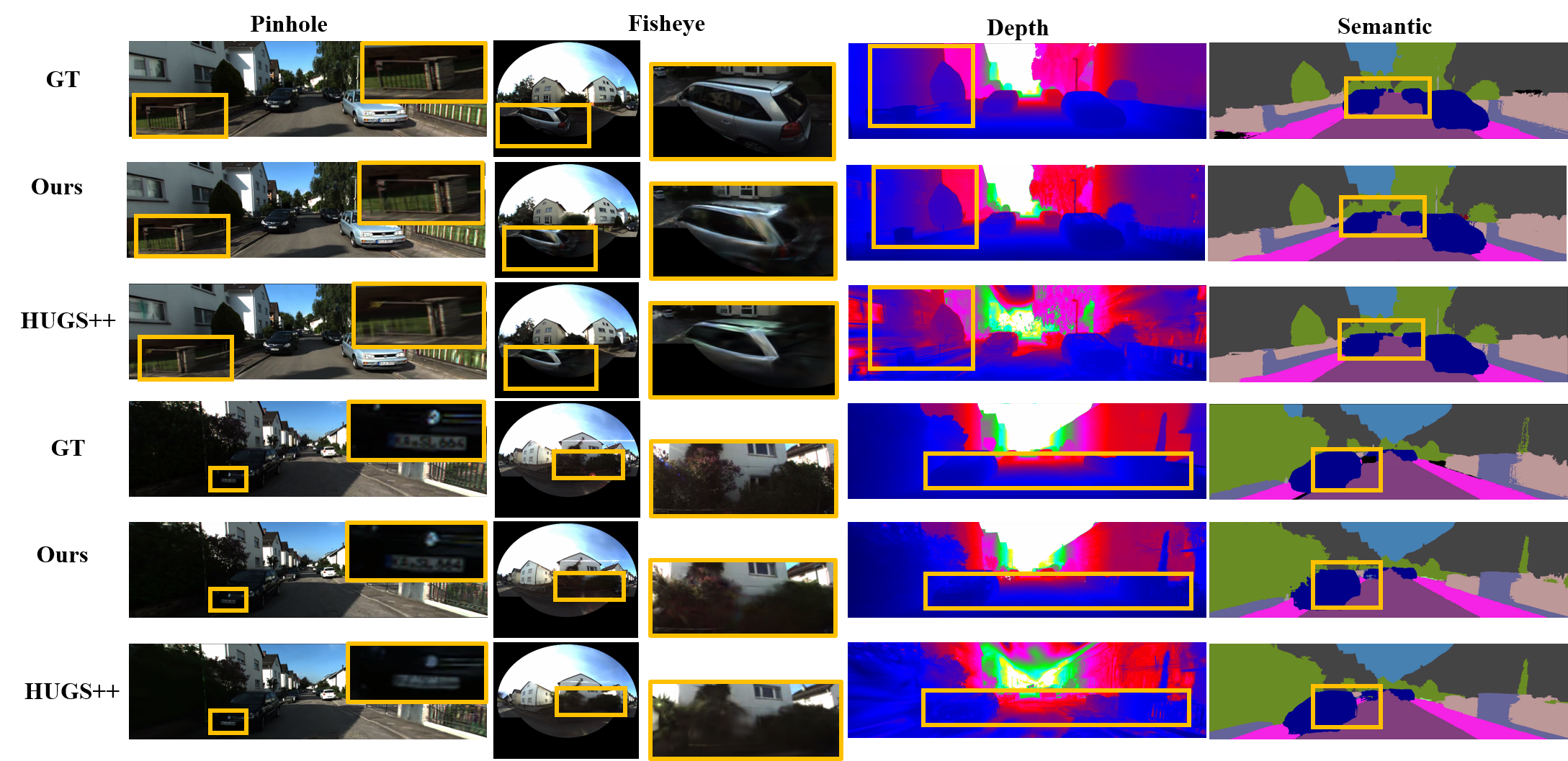}
  \vspace{-0.3cm}
  \caption{Rendering results of multiple camera model simulation on KITTI-360.
  We highlight the key differences in yellow boxes.}
  \label{fig:sota_multicam}
 \end{figure*}

\subsection{Driving Scene Fisheye Camera Simulation}
\label{sec:exp_fisheye_simulation}
In this experiment, we verify the effectiveness of the proposed fisheye rendering method on the KITTI-360 dataset~\cite{liao2022kitti}.
We compare the image quality of both global image and three representative local zones, namely, A, B, and C, as illustrated in \cref{fig:small_zone}.
Details of the experimental setup can be found in the supplementary material.

We present the quantitative results in \cref{tab:kitti360_res} and the qualitative results in \cref{fig:kitti_render_results}.
Overall, our approach achieves the best rendering quality both globally and locally (at zones A, B and C), while maintaining real-time rendering speed. 
In comparison, 3DGS+Undistort performs poorly in fisheye simulation, indicating that direct distortion is not a suitable choice.
Besides, from \cref{tab:kitti360_res} and \cref{fig:w_wo_stretching}, we can see that our method without stretching yields slightly worse image quality, especially for texture details, but achieves significantly faster rendering speed of 138 FPS.
Also, skipping stretching increases the number of Gaussians by $40\%$, which increases the risk of GPU memory overflow.
The reason is that small Gaussians are insensitive to changes in the distribution caused by light rays distortion and the optimizer tends to approximate the scene with a large number of small 3D Gaussians in exchange for better rendering quality.

\begin{table}[t!]
  \resizebox{0.47\textwidth}{!}{
    \centering
    \begin{tabular}{l|ccc|ccc}
        \toprule
        \multirow{2}{*}{Method}          & \multicolumn{3}{c|}{Pinhole} & \multicolumn{3}{c}{Fisheye} \\
                                                        & PSNR$\uparrow$ & SSIM$\uparrow$ & LPIPS$\downarrow$ & PSNR$\uparrow$ & SSIM$\uparrow$ & LPIPS$\downarrow$ \\
        \midrule       
        Instant-NGP\cite{muller2022instant}            & 24.6           & 0.808          & 0.181      & - & - & - \\
        UniSim-SF\cite{yang2023unisim,tao2024alignmif} & 24.9           & 0.812          & 0.184      & - & - & - \\
        AlignMiF\cite{tao2024alignmif}                 & 25.3           & 0.826          & 0.164      & - & - & - \\        
        Ours(pinhole)                                  & \underline{\textbf{25.7}}  & \underline{\textbf{0.881}} & \underline{\textbf{0.102}}      & - & - & - \\        
        \midrule
        % 3DGS\cite{kerbl3Dgaussians}+Undistort          &  \\
        HUGS++\cite{zhou2024hugs}                      & 25.2           & 0.866          & 0.146      & 24.9  & 0.879  & 0.241 \\        
        Ours                                           & \underline{\textbf{26.1}}  & \underline{\textbf{0.886}} & \underline{\textbf{0.099}}
                                                       & \underline{\textbf{26.2}}  & \underline{\textbf{0.897}} & \underline{\textbf{0.185}} \\
        \bottomrule
    \end{tabular}
  }
  \vspace{-0.3cm}
    \caption{Results of multiple camera model simulation on KITTI-360.
    ``HUGS++'' denotes using our method for additionally training fisheye cameras, and for a fair comparison, ``HUGS++'' uses the same initialization as our approach.
    Some results are borrowed from~\cite{tao2024alignmif}.
    }
    \label{tab:sota_multicam}
\end{table}

\subsection{Multiple Camera Model Simulation}
\label{sec:exp_multi_camera_simulation}

In this experiment, we evaluate our UniGaussian framework for driving scene simulation from multiple camera models.
We conduct our experiments on KITTI-360~\cite{liao2022kitti}, because it is a real-world autonomous driving dataset that provides both pinhole and fisheye images, while the other commonly used datasets provide only one type of image.

As shown in \cref{tab:sota_multicam}, for pinhole camera simulation, for fair comparison with AlignMiF and UniSim-SF, we also report the results of our approach with pinhole camera only.
We can see that both ours and ours(pinhole) achieve better results than the state-of-the-art driving scene reconstruction methods, such as HUGS, AlignMiF and UniSim-SF.
As for fisheye camera simulation, we add our fisheye rendering method to HUGS (named HUGS++) because HUGS does not support fisheye cameras.
The results in \cref{tab:sota_multicam} show that our approach outperforms the modified method.
Besides, as shown in \cref{fig:sota_multicam}, our approach generates better rendering images with more fine-grained details.
For example, our approach renders the license plate, the fence, the taillight, \etc, while HUGS++ generates worse rendering images.
Moreover, we also visualize some semantic and depth maps in \cref{fig:sota_multicam}.
We can see that our multimodal outputs achieve slightly better holistic 3D scene understanding.

\begin{table}[t!]
  \resizebox{0.47\textwidth}{!}{
    \centering
    \begin{tabular}{l|cc}
        \toprule
        Component         & Pinhole PSNR$\uparrow$ & Fisheye PSNR$\uparrow$ \\      
        \midrule
        Ours                                                  & \underline{\textbf{26.10}} & \underline{\textbf{26.19}} \\
        Ours w/o depth supervision         & 25.96     & \underline{\textbf{26.19}}  \\
        Ours w/o semantic supervision      & 26.01     & 25.85\\
        Ours w/o normal supervision        & 26.02     & 26.17\\
        Ours w/o adaptive control and $\mathcal{L}_{reg}$     & 25.76     & 25.75\\
        \bottomrule
    \end{tabular}
  }
  \vspace{-0.3cm}
    \caption{Ablation study on KITTI-360.
    }
    \label{tab:ablation_multicam}
\end{table}

\subsection{Ablation study}
\label{sec:exp_ablation_stuey}
The ablation studies of the stretching and approximation operations of the proposed rendering method are discussed in \cref{sec:exp_fisheye_error} and \cref{sec:exp_fisheye_simulation}.
In this experiment, we further evaluate other components of our approach.
From \cref{tab:ablation_multicam}, we can see that our approach with all components performs the best and the supervisions from different modalities, including depth, semantic, and normal, are helpful for model optimization.
These modalities also facilitate better holistic understanding of driving scenes as shown in \cref{fig:sota_multicam}.
Besides, from the last row of \cref{tab:ablation_multicam}, we can see that without the adaptive control and $\mathcal{L}_{reg}$, the PSNRs are slightly worse, indicating the importance of learning a compact representation in driving scene reconstruction.

\section{Conclusion}
\label{sec:conclusion}

In this work, we present a new differentiable rendering method of 3D Gaussians tailored for fisheye cameras and propose a new framework for learning unified 3D Gaussians of driving scenes from multiple camera models.
Our method enables holistic driving scene understanding by modeling multiple sensors and modalities.
Our experimental results on real-world autonomous driving dataset verify the effectiveness of the proposed method.

{
    \small
    \bibliographystyle{ieeenat_fullname}
    \bibliography{main}
}
\clearpage
\setcounter{page}{1}
\maketitlesupplementary

This supplementary material is organized as follows:
in \cref{sec:exp_setup}, we present more details of the experimental setups;
in \cref{sec:implementation_details}, we provide the implementation details of our approach;
in \cref{sec:com_fisheyegs}, we present the experimental comparison with Fisheye-GS~\cite{liao2024fisheye};
in \cref{sec:lane_shit}, we present results of novel view synthesis with lane shift;
in \cref{sec:lidar_sim}, we discuss the details of our approach to Lidar simulation;
in \cref{sec:discuss_nerf}, we discuss the key difference between NeRF-based methods and our approach;
in \cref{sec:limitation_future_work}, we discuss the limitation and future work of our approach.

\section{Experimental Setups}
\label{sec:exp_setup}

\subsection{Experimental Setup for Fisheye Rendering Geometric Error Analysis}
The pinhole and fisheye cameras are set to have the same FOV and resolution during rendering.
To ensure the rendering quality of the ``project-convert'' method, the FOV is set close to the FOV of the training data.
The resolution needs to be set to a large value because the pinhole image will shrink to the center area after distortion.
The 3DGS hyper-parameters are mainly set following~\cite{kerbl3Dgaussians}.

\subsection{Experimental Setup for Driving Scene Fisheye Camera Simulation}
\paragraph{Dataset.}
We conduct experiments on the KITTI-360 dataset~\cite{liao2022kitti}.
KITTI-360 contains images captured from fisheye cameras in real-world driving scenes and provides camera calibration and ego-vehicle pose.
We select a segment of 221 images (frame ids 227-447 from ``2013\_05\_28\_drive\_0000\_sync'') and evenly select every eighth image as the test set while the others are used as the training set.
In addition to comparing the global image quality, three local zones (A, B, C) are compared.
As shown in \cref{fig:small_zone}, zone ``A'' (Blue) is the bottom area where objects appear usually very close to the camera, zone ``B'' (Green) is the sky with little texture where the geometry is not very accurate, and zone ``C'' (Red) is far away from the camera with the large distortion in this region. 

\paragraph{Camera Model Conversion.}
The MEI camera model is the only model provided for the fisheye camera on KITTI-360, so to evaluate our method on both the MEI and Kannala-Brandt models, we convert the MEI model to the Kannala-Brandt model.
Specifically, the conversion is an approximation of $\theta_d(\theta)$ determined by Eq.~\eqref{eq:MEI_theta_d} of the main paper with the series expansion Eq.~\eqref{eq:Kannala-Brandt} of the main paper.
Note that when $\theta=0$, the first derivative of Eq.~\eqref{eq:Kannala-Brandt} equals $1$, while the first derivative of Eq.~\eqref{eq:MEI_theta_d} is $1/(1+\xi)$.
To eliminate this difference, Eq.~\eqref{eq:MEI_theta_d} is modified as:
\begin{small}
\begin{equation}
    \theta_d = \arctan r_d = \arctan \left( \left( \chi + k_1 \chi^3 + k_2 \chi^5 \right) (1+\xi) \right).    
    \label{eq:MEI_theta_d_modified}
\end{equation}
\end{small}%
Correspondingly, the focal length of Kannala-Brandt is set as $f_x=\gamma_1/(1+\xi)$ and $f_y=\gamma_2/(1+\xi)$.
The first step of the conversion is to generate a set of $\theta$ and $\theta_d$ by using Eq.~\eqref{eq:MEI_theta_d_modified},
where $\theta \in(0,\pi/2]$.
Then, $k_i$(i=1,2,3,4) of the Kannala-Brandt model can be estimated by the least squares method.
In our experiments, we test both the MEI and Kannala-Brandt camera models but the results are almost the same, so we by default report the results with the MEI model.

\paragraph{Compared Methods.}
We compare our approach with three NeRF-based methods, including Instant-NGP~\cite{muller2022instant}, Nerfacto-big~\cite{nerfstudio}, Zip-NeRF~\cite{barron2023zip}, and one 3DGS-based method, namely ``3DGS~\cite{kerbl3Dgaussians}+Undistort''.
To adapt 3DGS for driving scene reconstruction with fisheye images, we rectangularize fisheye images for reconstruction and render scene with the pinhole camera model and distort to fisheye images (named 3DGS+Undistort).
Besides, the NeRF-based methods are implemented based on~\cite{nerfstudio}.

\subsection{Experimental Setup for Multiple Camera Model Simulation}
\paragraph{Dataset.}
We conduct our experiments on KITTI-360~\cite{liao2022kitti}, because it is a real-world autonomous driving dataset that provides both pinhole and fisheye images, while the other commonly used datasets provide only one type of image.
For evaluation, we select four sequences as~\cite{tao2024alignmif}.
Each sequence contains 64 frames, and we select every fourth frame as the test set while the others are used as the training set.
Note that although there are fisheye datasets for autonomous driving, such as Woodscape~\cite{yogamani2019woodscape}, they are constructed for perception tasks (\eg, object detection, segmentation, \etc) rather than driving scene reconstruction, so they often only have discrete frames and lack continuous video sequence data.
In contrast, KITTI-360 provides a complete real-world autonomous driving sensor suite, including fisheye, pinhole and LiDAR, so evaluation on KITTI-360 for multiple camera model simulation is comprehensive to verify the effectiveness of our approach.

\begin{table*}[t]
  % \resizebox{0.3\textwidth}{!}{
    \centering
    \begin{tabular}{l|cccc|c}
        \toprule
        Methods                             & Alameda & Berlin & London  & Nyc    & Average \\      
        \midrule
        Ours                                & \underline{\textbf{26.0}}    & 23.8   &  \underline{\textbf{27.9}}   & \underline{\textbf{22.3}}   & \underline{\textbf{25.0}}  \\  
        Fisheye-GS~\cite{liao2024fisheye}   & 24.1    & \underline{\textbf{24.2}}   &  25.1   & 20.3   & 23.4  \\  
        \bottomrule
    \end{tabular}
  % }
  \vspace{-0.3cm}
    \caption{Comparison with Fisheye-GS on the Zip-NeRF(fisheye) dataset.
    Results are in terms of PSNR$\uparrow$.
    }
    \label{tab:com_fisheyegs}
\end{table*}

\paragraph{Compared Methods.}
To the best of our knowledge, no existing driving scene reconstruction method simulates both pinhole and fisheye cameras in a unified framework.
HUGS~\cite{zhou2024hugs} is a state-of-the-art 3DGS-based method for driving scene reconstruction and 3D scene understanding.
We therefore use it as a baseline method for comparison.
We modify HUGS based on the official code and the paper and additionally use our differentiable rendering method to train HUGS for fisheye camera simulation.
For the other compared driving scene simulation methods, we borrow the results from~\cite{tao2024alignmif}.
Note that although there are some other driving scene simulation methods, they are mostly designed for pinhole camera simulation, so we choose the representative HUGS~\cite{zhou2024hugs}, AlignMiF\cite{tao2024alignmif} and UniSim-SF\cite{yang2023unisim,tao2024alignmif} for comparison in our experiments.
Modifying existing state-of-the-art methods for fisheye camera simulation in autonomous driving is beyond the scope of this work.

\section{Implementation Details}
\label{sec:implementation_details}
We implement our approach using python and pytorch.
Following~\cite{kerbl3Dgaussians,zhou2024hugs}, we set the initial position learning rate to $1.6\times10^{-4}$ with a decay to $1.6\times10^{-6}$ at the last iteration.
Both scaling and rotation learning rates are set to $0.001$, the learning rate of the spherical harmonics feature is set to $0.0025$ and the opacity learning rate is set to $0.05$.
Adaptive Density Control is performed every 100 iterations starting from 500 iterations.
For composite scene Gaussians, the background Gaussians are initialized with LiDAR point clouds, the dynamic Gaussians are randomly initialized, and the sky Gaussians are uniformly initialized in a distant (more than 100 meters) spherical region for simplicity, but more advanced sky cubemaps~\cite{yan2024street} can be used.
The training loss $\mathcal{L}$ of our approach is defined in Eq.~\eqref{eq:loss} of the main paper.
The details of these losses and regularization terms are listed below.

\paragraph{Image Losses $\mathcal{L}_{rgb}^{P}$ and $\mathcal{L}_{rgb}^{F}$.} They are the reconstruction losses between the ground-truth and the rendering pinhole/fisheye images, which are defined as:
\begin{equation}
  \mathcal{L}_{rgb} = (1-\lambda_{rgb})\mathcal{L}_1 + \lambda_{rgb}\mathcal{L}_{SSIM},
\end{equation}
where $\mathcal{L}_1$ is the L1 loss, $\mathcal{L}_{SSIM}$ is a D-SSIM term~\cite{kerbl3Dgaussians}, and $\lambda_{rgb}$ is set to 0.2 following~\cite{kerbl3Dgaussians}.

\paragraph{Depth Loss $\mathcal{L}_{d}$.} It is the depth loss computed between the rendering depth and the monocular depth $\mathcal{L}_{d-mono}$ or LiDAR depth $\mathcal{L}_{d-lidar}$.
Although the depth derived from LiDAR is accurate, it can only supervise the masked region with the projected point clouds.
On the other hand, the monocular depth is coarse but can supervise the whole depth map.
Thus, we define $\mathcal{L}_{d}$ as:
\begin{equation}
  \mathcal{L}_{d} = \mathcal{L}_{d-lidar} + \mathcal{L}_{d-mono},
\end{equation}
where $\mathcal{L}_{d-lidar}$ is the L1 loss between the depth derived from LiDAR point clouds and the masked region from the rendered depth map, and $\mathcal{L}_{d-mono}$ is the Pearson depth loss~\cite{xiong2023sparsegs} between the rendered depth map and the monocular depth map computed with~\cite{hu2024metric3d}.

\paragraph{Semantic Loss $\mathcal{L}_{s}$.} It is the semantic loss for the rendering semantic map $M_s$ and the predefined 2D semantic segmentation map~\cite{Liao2022PAMI}. This is implemented with the cross-entropy loss to classify the semantic logits and its weight is set to 0.01. This loss helps to generate semantic maps for driving scenes, improving the holistic driving scene understanding.

\paragraph{Normal Loss $\mathcal{L}_{n}$} It is the normal consistency loss regularizing the rendering normal $N_p$ and the normal derived from the depth $N_d$. It encourages a better geometric representation of the driving scene. We define it as:
\begin{equation}
  \mathcal{L}_{n} = \mathcal{F}_m(\|1-{N_p^T}N_d\|_1),
\end{equation}
where $\mathcal{F}_m$ denotes the mean operation.

\paragraph{Opacity and Scale Regularization Term $\mathcal{L}_{reg}$.}
Following~\cite{kheradmand20243d}, we employ the Gaussian opacity and scale regularization term to encourage a compact scene Gaussian representation. It is defined as:
\begin{equation}
  \mathcal{L}_{reg} = \lambda_{reg}\left( \mathcal{F}_m(\left|o \right|)+\mathcal{F}_m(\left|s \right|) \right),
\end{equation}
where $o$ are the Gaussian opacities, $s$ are the Gaussian scales and $\lambda_{reg}$ is set to 0.01.

\begin{figure*}[!t] 
  \centering
  \includegraphics[width=0.95\textwidth]{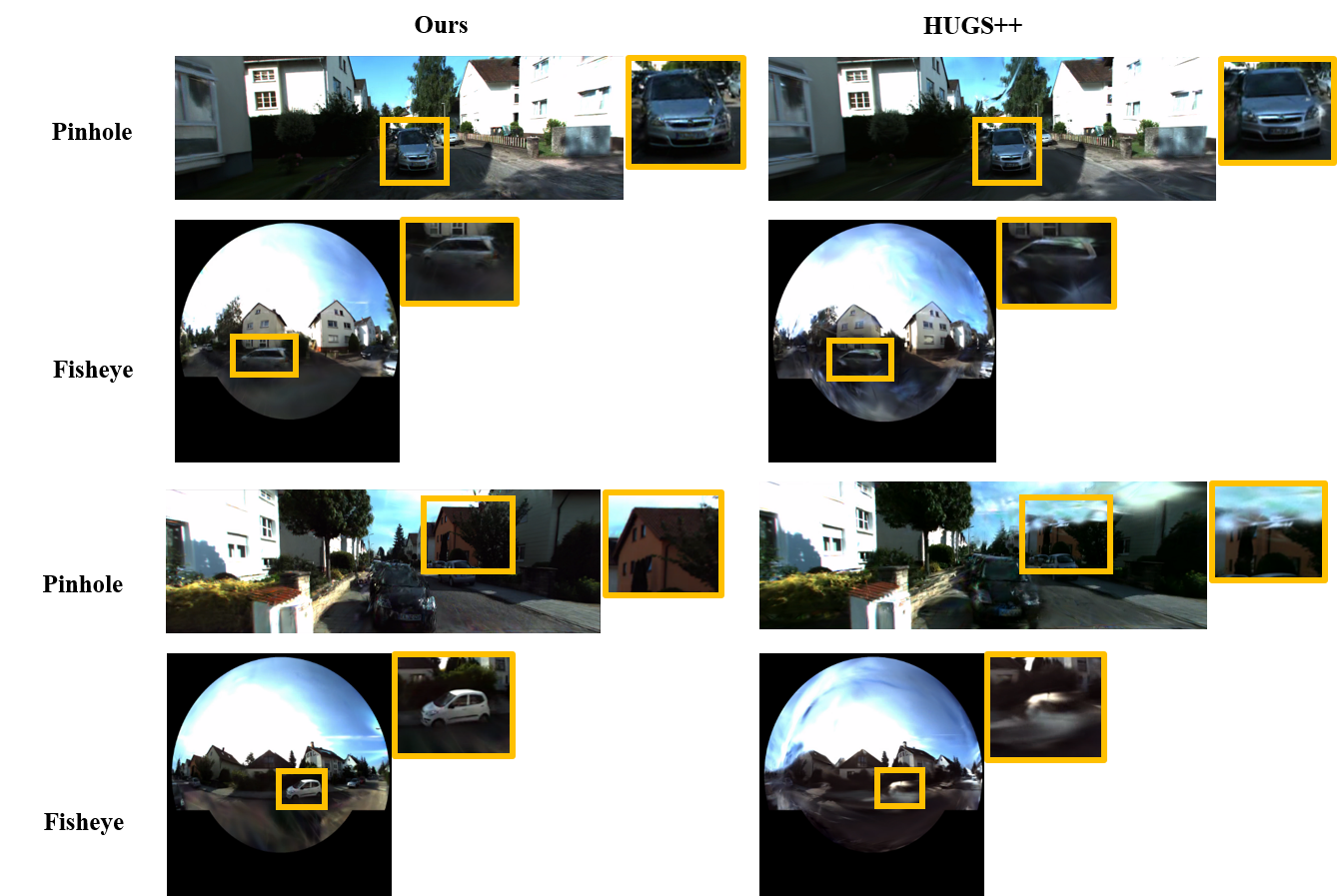}
  \vspace{-0.3cm}
  \caption{Qualitative results of novel view synthesis with lane shift on KITTI-360.}
  \label{fig:lane_shift}
 \end{figure*}

\section{Experimental Comparison with Fisheye-GS}
\label{sec:com_fisheyegs}
As discussed in \cref{sec:related_work_3dgs}, Fisheye-GS~\cite{liao2024fisheye} is only based on ideal camera models with equidistant projection, which hinders its use in driving scene reconstruction because fisheye cameras in driving scene are usually generic models and have large FOVs.
Since Fisheye-GS is a state-of-the-art method adapting 3DGS to fisheye cameras, it would be interesting to compare our differentiable fisheye rendering method with Fisheye-GS.
To this end, we conduct experimental comparison on Zip-NeRF(fisheye)~\cite{barron2023zip} with four large scenes, namely, Alameda, Berlin, London, and Nyc.
These scenes are respectively captured with a fisheye lens of 180 degree.
We select every eighth frame as the test set while the others are used as the training set.
As shown in \cref{tab:com_fisheyegs}, our approach achieves better performance compared with Fisheye-GS.
Specifically, in the Alameda, London and Nyc scenes, our approach achieves significantly better results, while in the Berlin scene, our approach achieves comparable performance compared with Fisheye-GS.
On average, our approach yields PSNR of 25.0 dB while Fisheye-GS obtains PSNR of 23.4 dB.
These results verify the superiority of our approach over Fisheye-GS.

\begin{table}[t!]
  % \resizebox{0.3\textwidth}{!}{
    \centering
    \begin{tabular}{l|cc}
        \toprule
        Methods        & Pinhole FID$\downarrow$@3m & Fisheye FID$\downarrow$@3m \\      
        \midrule
        Ours           & 182.7                   & 193.7 \\  
        HUGS++         & 191.7                   & 300.8 \\
        \bottomrule
    \end{tabular}
  % }
  \vspace{-0.3cm}
    \caption{Quantitative results of novel view synthesis with lane shift on KITTI-360.
    }
    \label{tab:lane_shift}
\end{table}

\section{Novel View Synthesis with Lane Shift}
\label{sec:lane_shit}
In driving scene reconstruction, it is important to synthesize novel views after the lane shift (left or right) of the ego vehicle for real-world simulators.
In this experiment, we further present the results of novel view synthesis with lane shift @ 3 meters of the ego vehicle on KITTI-360.
From \cref{tab:lane_shift} and \cref{fig:lane_shift}, we can see that our approach is able to synthesize high-quality rendered images for novel view synthesis with lane shift while HUGS++ generates significantly worse results.
For example, the FID of our approach is significantly better than HUGS++ and the buildings and vehicles in the rendered images of our approach are much clearer and more complete.

\begin{figure*}[!t] 
  \centering
  \includegraphics[width=0.7\textwidth]{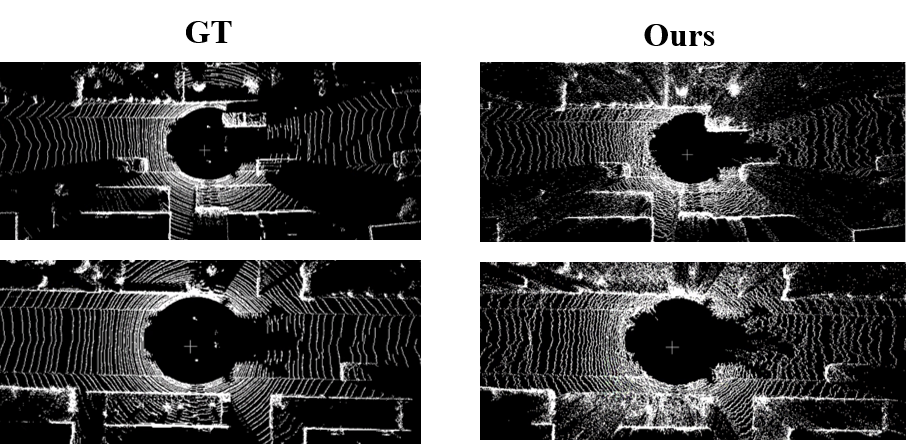}
  \vspace{-0.3cm}
  \caption{Visualization of simulated LiDAR point clouds on KITTI-360.}
  \label{fig:lidar_sim}
 \end{figure*}

\section{Details of the Optional LiDAR Simulation}
\label{sec:lidar_sim}
In this section, we discuss the potential of our approach to LiDAR point clouds simulation.
As mentioned in the main paper, our unified framework can be extended to simulate point clouds by extracting points from the rendering depth maps.
This is achieved by predefining the LiDAR scans based on real-world LiDAR parameters and mapping the points of the scans from the world coordinate space to the camera coordinate space.
By default, depth maps are associated with the pinhole camera coordinate space, but KITTI-360 only has two front pinhole cameras with narrow FOVs so the rendering depth maps cannot cover all point clouds.
To solve this problem, we further render depth maps from eight pseudo cameras, covering the left, right, front and back directions, as well as downward towards the road from each of these directions.
In this way, we extract each point cloud from the nearest depth map.
Furthermore, to simulate the intensities of point clouds, we render intensity maps via $\alpha$-blending of the 3D logits of 3D Gaussians in the rasterizer.
Then, the intensities of point clouds are extracted at the corresponding locations on the intensity maps.
To facilitate model optimization for LiDAR simulation, a LiDAR loss $\mathcal{L}_{l}$ is added to Eq.~\eqref{eq:loss} of the main paper, which is defined as:
\begin{equation}
  \mathcal{L}_{l} = \lambda_{l}\left( \mathcal{F}_m(\left|x_{sim} - x_{gt}\right|)+\mathcal{F}_m(\left|I_{sim} - I_{gt} \right|) \right),
\end{equation}
where $x_{sim}$ and $I_{sim}$ are the simulated point cloud positions and intensities, $x_{gt}$ and $I_{gt}$ are ground truths, and $\lambda_{l}$ is set to $0.1$.
In \cref{fig:lidar_sim}, we visualize some simulated LiDAR point cloud maps and intensity maps.
From these results, we can observe that the simulated point cloud and intensity maps are close to the real LiDAR, especially for the nearby regions.
On the other hand, we can also observe some noisy points because the indirect simulation strategy relies heavily on the accuracy of the depth maps.
Note that the LiDAR simulation is not the focus of this work and more effort should be made in order to improve LiDAR simulation with 3DGS and to simulate more complex LiDAR phenomena.

\section{Key Difference between NeRF-based methods and our approach}
\label{sec:discuss_nerf}
Although there have been some NeRF-based methods for fisheye rendering, such as~\cite{choi2024dico}, they are mostly time-consuming and cannot be combined with Gaussian splatting for driving scene simulation.
Besides, in~\cite{choi2024dico}, Choi~\etal show that the performance of their approach is close to Instant-NGP and NeRFacto, while our experiments in Tab.~\ref{tab:kitti360_res} show that our approach significantly outperforms Instant-NGP and NeRFacto.
More importantly, we did not thoroughly compare these NeRF-based methods because the novelty of our approach is a new fisheye method to solve limitation of 3DGS and a unified framework for driving scene reconstruction, while the performance of NeRF-based methods does not diminish the novelty of our approach.

\section{Limitation and Future Work}
\label{sec:limitation_future_work}

Due to the nature of explicit 3D Gaussian representations, 3DGS may not provide sufficient details when the viewing distance is close to the observation areas.
This issue may be exacerbated when adapting 3DGS to fisheye cameras in driving scene reconstruction due to the capability of fisheye cameras to observe nearby vehicles and buildings along the road.
However, our experiments in \cref{sec:exp_fisheye_simulation} show that even for zone ``A'' (Blue) in \cref{fig:small_zone}, \ie, the bottom area where objects appear usually very close to the camera, our approach achieves better reconstruction and renders better results compared with existing methods.
To further address this limitation, future endeavors can focus on dynamically adjusting the variable lower bound of scale for 3D Gaussians observed at the close proximity based on the anticipated nearest observation distance.
Such an approach would enhance the adaptability and efficiency of the rendering process, capturing nearby details.
Besides, our future work also aims to develop a real-world autonomous driving simulator.

\end{document}